\documentclass[]{aidata}

\usepackage[utf8]{inputenc}
\usepackage{nicefrac}
\usepackage{amsmath}
\usepackage{amssymb}
\usepackage{amsfonts}
\usepackage{array}
\usepackage{makecell}
\usepackage{tabularx}
\usepackage{longtable}
\usepackage{wrapfig}
\usepackage{enumitem}
\usepackage{colortbl}   
\usepackage{pifont}     

\definecolor{colorAgent}{HTML}{F3E5F5}
\definecolor{colorBaseline}{HTML}{F5F5F5}
\definecolor{colorGT}{HTML}{E3F2FD}
\definecolor{colorUp}{HTML}{2E7D32}
\definecolor{colorDown}{HTML}{C62828}
\definecolor{checkgreen}{HTML}{2b8a3e}
\definecolor{crossred}{HTML}{c92a2a}
\definecolor{partialorange}{HTML}{e67e22}
\definecolor{ourrow}{HTML}{FFF9DB}

\title{Before the Action: Benchmarking LLMs on Prospective Hypothesis Discovery}

\author[1,2,*]{Tianyun Zhong}
\author[1,2,*]{Wangyi Jiang}
\author[3]{Wei Wang}
\author[2]{Xuanang Chen}
\author[2]{Yaojie Lu}
\author[1]{Shiwei Ye}
\author[3]{Yuzhen Shi}
\author[3]{Boyu Yang}
\author[3]{Jinghang Wang}
\author[3]{Han Li}
\author[3]{Weiqi Zhai}
\author[3]{Bing Zhao}
\author[3,\dagger]{Hu Wei}
\author[3]{Haiyang Yu}
\author[3,\dagger]{Yongbin Li}
\author[2]{Hongyu Lin}
\author[2]{Le Sun}
\author[2,\dagger]{Xianpei Han}

\affiliation[1]{University of Chinese Academy of Sciences}
\affiliation[2]{Institute of Software, Chinese Academy of Sciences}
\affiliation[3]{Alibaba Group}

\contribution[*]{Equal contribution. Work done during internship at Alibaba Group.\\
\email{zhongtianyun2023@iscas.ac.cn}, \email{jiangwangyi2020@iscas.ac.cn}\\   
${\dagger}$Corresponding authors.}  
      
\abstract{

Large language models (LLMs) excel at answering given questions, yet their ability to navigate open-ended, pre-conclusion discovery remains largely unmeasured. We introduce \textit{Prospective Hypothesis Discovery (PHD)}, which asks models to autonomously construct grounded, discriminative, and testable hypothesis spaces from inconclusive evidence, including anomalous observations and fragmented records, to guide subsequent investigation. To evaluate this, we present \textsc{HypoArena}, comprising \textsc{HypoData} (988 cases across six domains) and \textsc{HypoEval} (an evaluation framework for open-ended hypotheses). We construct \textsc{HypoData} via \textit{Retrospective Context Regression}, a pipeline that reconstructs pre-conclusion contexts from expert documents by removing explicit conclusions while preserving factual substrates. Since PHD admits multiple valid outputs, \textsc{HypoEval} combines bidirectional pairwise judgments with Bradley--Terry--Davidson aggregation for ranking, alongside six-dimensional rubric scoring for diagnosis. Experiments on 15 frontier LLMs reveal clear capability stratification and show that arena evaluation resolves finer-grained model differences while aligning strongly with human expert judgment. Together, these results support treating PHD as a distinct target for evaluating the investigative reasoning of LLMs. All code, data, and evaluation tools are publicly available at \href{https://github.com/SKYLENAGE-AI/HypoArena}{\texttt{github.com/SKYLENAGE-AI/HypoArena}} and \href{https://huggingface.co/datasets/HypoArena/HypoData}{\texttt{huggingface.co/datasets/HypoArena/HypoData}}.

}

\begin{document}

\maketitle

\section{Introduction}

\begin{figure}
    \centering
    \includegraphics[width=0.9\linewidth]{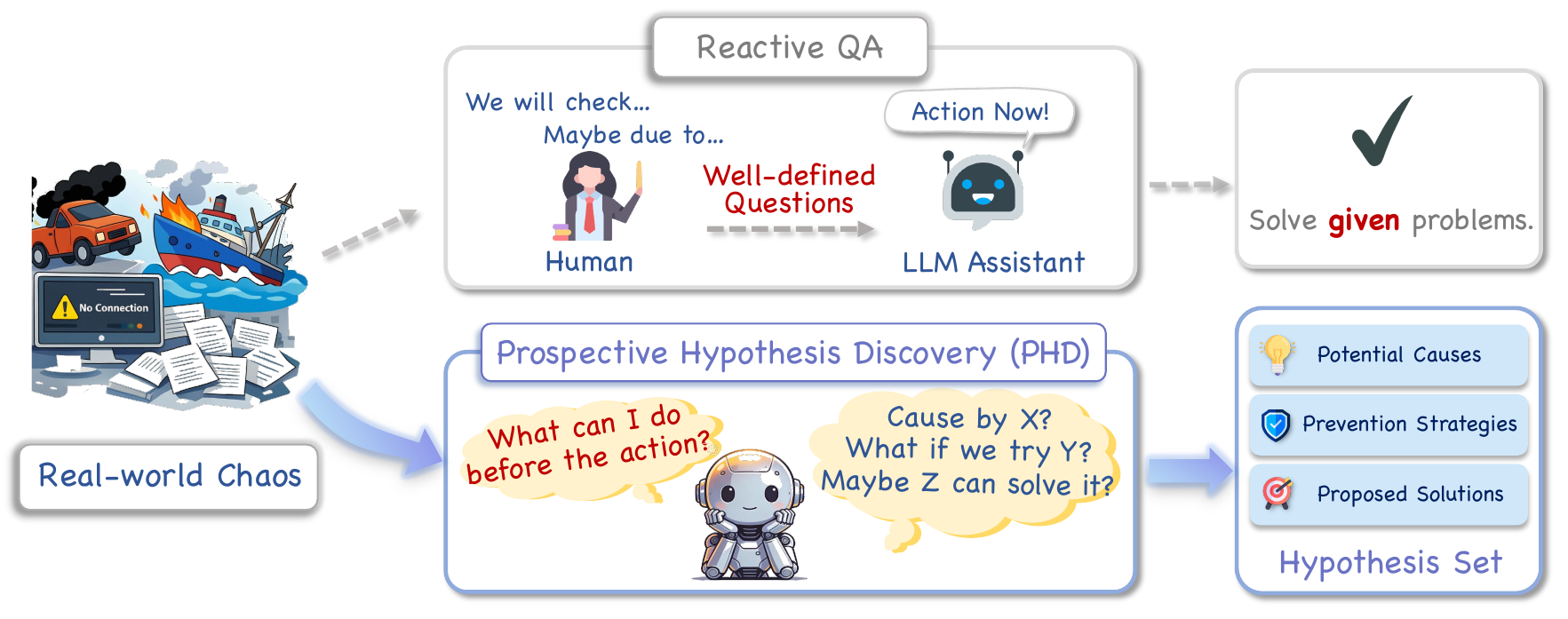}
    \caption{\textbf{From reactive QA to prospective hypothesis discovery.} Conventional QA hands the model a pre-formed question (\emph{reactive}). \textsc{HypoArena} instead reconstructs a real-world context in pre-conclusion form and asks the model to \emph{proactively} construct a plausible hypothesis space---the capability we term \textit{Prospective Hypothesis Discovery} (PHD).}
    \label{fig:overview}
\end{figure}

Real-world discovery rarely starts from a well-posed question. In scientific research, accident investigation, and financial analysis, the first encounter with a problem is typically a tangle of anomalous observations, fragmented facts, and incomplete records. The defining challenge at this stage is not retrieving a known answer but constructing a plausible hypothesis space: a set of grounded, verifiable, and mutually discriminative claims worth investigating next. We call this capability \textit{Prospective Hypothesis Discovery} (PHD), and argue that it is a distinct and largely unmeasured competence of LLMs. For instance, given sudden server errors and fragmented logs, a capable system should propose several verifiable directions such as memory leaks or microservice deadlocks before any cause has been confirmed. Figure~\ref{fig:overview} contrasts this setting with conventional QA-style evaluation.


Existing benchmarks do not directly target this capability. QA and agent benchmarks~\citep{egg2025dabstep,jingdsbench,mitchener2025kosmos} score models against predefined goals or task-completion criteria. Scientific discovery and reasoning benchmarks~\citep{discovery-bench,insight-bench,hypo-bench,sci-arena,inno-eval} typically supply structured data or pre-formulated research questions. Open-ended idea-generation benchmarks~\citep{wang2026fire,idea-bench,ai-idea-bench,research-bench} start from paper abstracts or already-refined research backgrounds. None of these settings asks whether a model, given raw heterogeneous facts and no formed conclusion, can identify several plausible hypotheses, distinguish among them, and argue for each. A benchmark that isolates this question is missing.



Building such a benchmark is hard for two reasons, one on the data side and one on the metric side. On the data side, expert documents such as scientific papers, accident reports, and analyst notes are written after the conclusion is reached. They carry final claims, hypothesis statements, and explicit causal links, all of which are strong suggestive cues. Feeding such documents to a model collapses PHD into restating the known answer, while authoring conclusion-free contexts by hand is expensive and does not scale. On the metric side, a high-quality PHD output is not a single reference but an open hypothesis space~\citep{sican}. The same context typically supports multiple legitimate directions, so reference-matching metrics underestimate plausible hypotheses that happen to fall outside the gold set~\citep{research-bench,papineni2002bleu,lin2004rouge}, while absolute rubric scoring struggles to discriminate among open-ended outputs of comparable surface quality~\citep{liu2023g}.


To address these challenges, we construct \textsc{HypoArena}, a benchmark comprising 988 cases across six domains: Biomedical Science, Machine Learning, Social Science, Financial Analysis, IT Operations, and Safety Investigation. Each case is derived from human-authored documents such as research papers, investigation reports, or professional blogs. Our core motivation stems from the observation that when humans author these reports, they must generate plausible hypotheses, yet these intermediate hypotheses often remain unrecorded in the final artifacts. To approximate this intermediate reasoning stage, we construct our cases through Retrospective Context Regression. Rather than recovering the exact historical information state, this process reconstructs a model-visible context by preserving temporally admissible factual content while filtering explicit conclusions, target hypotheses, and retrospective causal attributions. We operationally refer to contexts satisfying these criteria as ``conclusion-free.'' The corresponding hypotheses and evidence are kept on the reference side for verification and diagnostic evaluation. Table~\ref{tab:benchmark_comparison_wide} positions \textsc{HypoArena} relative to prior benchmarks in scientific discovery and idea generation.

\begin{table}[t] 
\centering

\caption{Comparison of \textsc{HypoArena} with related benchmarks. We categorize existing benchmarks into three groups: QA \& Agents, Scientific Discovery, and Idea Generation. Here, D denotes database and Q denotes query.}
\footnotesize
\label{tab:benchmark_comparison_wide}
\renewcommand{\arraystretch}{1} 
\setlength{\tabcolsep}{6pt} 

\begin{tabular}{llc|ccc|c}
\toprule
\textbf{Benchmark} & \textbf{Input} & \textbf{Output} & \textbf{Retro.} & \textbf{Open} & \textbf{Multi-} & \textbf{Scale} \\
 & & & \textbf{Context} & \textbf{Hypo.} & \textbf{domain} & \\
\midrule

\rowcolor{gray!10} \multicolumn{7}{l}{\textit{Category 1: QA \& Agents}} \\
DABstep~\citep{egg2025dabstep} & Q + Data & Str/Num/List & \textcolor{crossred}{$\times$} & \textcolor{crossred}{$\times$} & \textcolor{crossred}{$\times$} & 450 \\
DSBENCH~\citep{jingdsbench} & Q + Data & Code / Num & \textcolor{crossred}{$\times$} & \textcolor{crossred}{$\times$} & \textcolor{crossred}{$\times$} & 540 \\

\midrule
\rowcolor{gray!10} \multicolumn{7}{l}{\textit{Category 2: Sci. Discovery \& Reasoning}} \\
DiscoveryBench~\citep{discovery-bench} & D + Goal & Single Hypo. & \textcolor{crossred}{$\times$} & \textcolor{crossred}{$\times$} & \textcolor{crossred}{$\times$} & 1.1k \\
InsightBench~\citep{insight-bench} & D + Goal & Insights & \textcolor{crossred}{$\times$} & \textcolor{crossred}{$\times$} & \textcolor{crossred}{$\times$} & 100 \\
HypoBench~\citep{hypo-bench} & Datasets & Single Hypo. & \textcolor{crossred}{$\times$} & \textcolor{crossred}{$\times$} & \textcolor{checkgreen}{$\checkmark$} & 194 \\
SciArena~\citep{sci-arena} & Q + Docs & Lit. Response & \textcolor{crossred}{$\times$} & \textcolor{checkgreen}{$\checkmark$} & \textcolor{checkgreen}{$\checkmark$} & 20K \\
InnoEval~\citep{inno-eval} & Idea & Eval Report & \textcolor{checkgreen}{$\checkmark$} & \textcolor{crossred}{$\times$} & \textcolor{crossred}{$\times$} & 761 \\

\midrule
\rowcolor{gray!10} \multicolumn{7}{l}{\textit{Category 3: Idea Generation}} \\
FIRE-Bench~\citep{wang2026fire} & Q & Full Research & \textcolor{checkgreen}{$\checkmark$} & \textcolor{checkgreen}{$\checkmark$} & \textcolor{crossred}{$\times$} & 30 \\
IdeaBench~\citep{idea-bench} & Abs & Idea & \textcolor{crossred}{$\times$} & \textcolor{checkgreen}{$\checkmark$} & \textcolor{crossred}{$\times$} & 2.4K \\
AI Idea Bench & Topic+Papers & Idea & \textcolor{crossred}{$\times$} & \textcolor{checkgreen}{$\checkmark$} & \textcolor{crossred}{$\times$} & 3.5K \\
ResearchBench~\citep{research-bench} & Q + Survey & Inspiration & \textcolor{crossred}{$\times$} & \textcolor{checkgreen}{$\checkmark$}  & \textcolor{checkgreen}{$\checkmark$} & 1.4K \\

\midrule
\rowcolor{colorGT} \textbf{HypoArena (Ours)} & \textbf{Context} & \textbf{Open Hypos} & \textcolor{checkgreen}{$\checkmark$} & \textcolor{checkgreen}{$\checkmark$} & \textcolor{checkgreen}{$\checkmark$} & \textbf{988} \\
\bottomrule
\end{tabular}
\end{table}


Since a single ``conclusion-free context'' often supports multiple plausible exploration directions, high-quality output is no longer a single answer but an open hypothesis space. To address issues such as the underestimation of innovative hypotheses and scoring biases inherent in exact match or rubric scoring, we design \textsc{HypoEval}, an arena-style evaluation protocol based on pairwise comparison. Under the same context, \textsc{HypoEval} moves away from a reliance on a single reference answer. Instead, it compares the hypothesis sets generated by two models to judge which side performs better across six dimensions: Contextual Grounding, Inferential Insight, Evidential Justification, Hypothesis-Space Breadth, Directional Distinctness, and Analytical Utility. Subsequently, we incorporate the Bradley--Terry--Davidson model to aggregate these pairwise judgments and calculate a more robust model ranking. Meanwhile, we retain rubric dimensions as diagnostic tools for analyzing model performance along the same six dimensions.

We conduct systematic empirical evaluations on 15 contemporary language models across six domains. Our \textsc{HypoEval} pairwise evaluation protocol addresses drawbacks of conventional rubric-based scoring. Across domains, arena evaluation produces a clearly stratified leaderboard, while rubric scores remain concentrated within sub-one-point bands on the 1--5 scale. Agent-mode structured analytic skills produce heterogeneous and model-dependent effects. We further assess the reliability of the arena rankings through human-expert alignment and cross-judge triangulation.



Our contributions are:

\begin{itemize}[leftmargin=1.5em, nosep]
    \item \textbf{Task Definition \& Benchmark}: We define the task of \textit{Prospective Hypothesis Discovery} (PHD) and introduce \textsc{HypoArena}, a benchmark of 988 cases across six scientific and analytical domains.

    \item \textbf{\textsc{HypoData} and Retrospective Context Regression}: We propose a scalable Forge--Audit construction method that reconstructs pre-conclusion contexts from completed expert documents under explicit leakage, temporal, faithfulness, and supportability controls.

    \item \textbf{\textsc{HypoEval} and Systematic Empirical Analysis}: We propose a dual-track evaluation framework that pairs an arena protocol with the Bradley--Terry--Davidson model as the primary ranking, complemented by rubric scoring as diagnostics. A study of 15 contemporary LLMs shows that arena evaluation provides finer-grained separation than rubric scoring and reveals protocol-dependent differences in model ordering.
\end{itemize}


\section{\textsc{HypoData}: Benchmarking Discovery}

\label{sec:hypodata}

\begin{figure}
    \centering
    \includegraphics[width=1\linewidth]{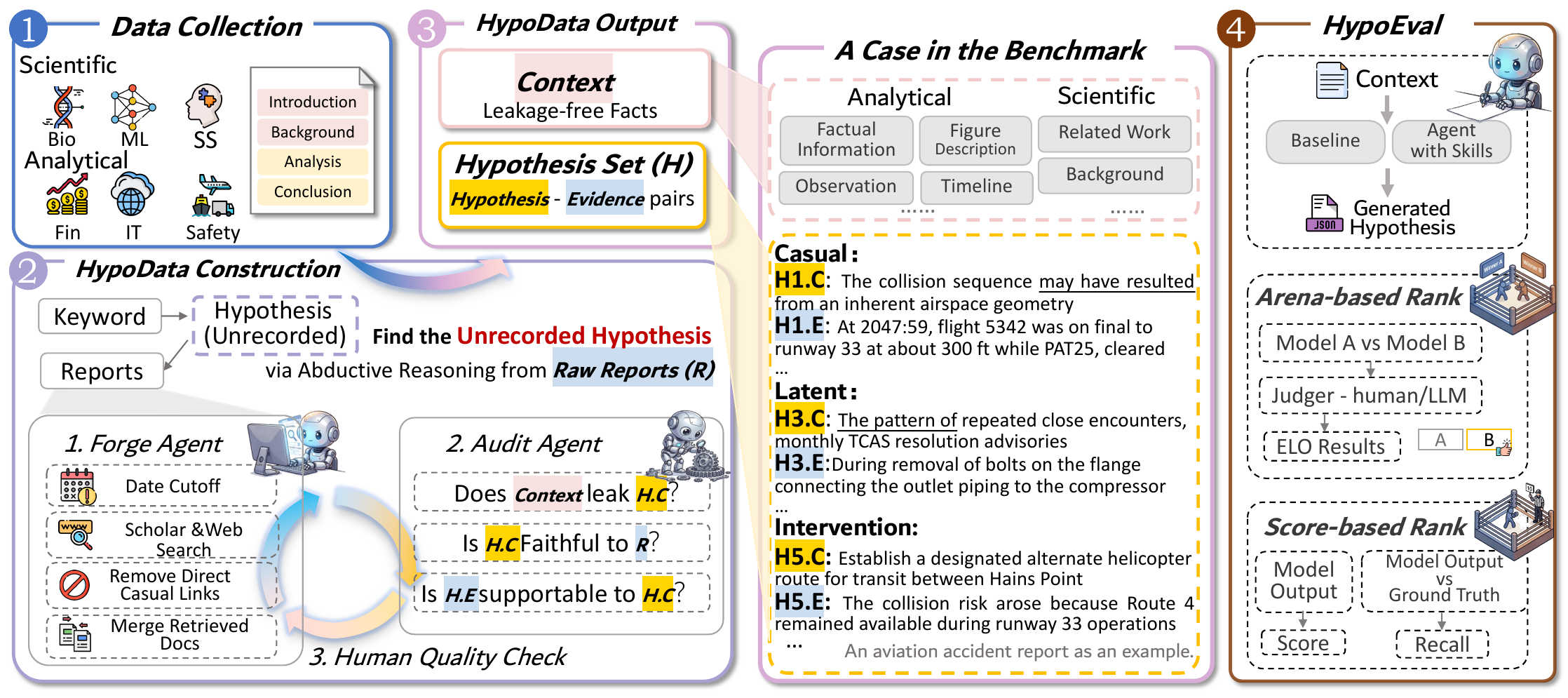}
    \caption{The \textsc{HypoArena} pipeline. (1) Multi-domain data collection from human-authored documents; (2) \textsc{HypoData}: Retrospective Context Regression rolls source documents back to a pre-conclusion factual state and extracts hidden hypothesis-evidence pairs as the reference; (3) Structured Context-Hypothesis-Evidence output, with the Context model-visible and the rest held out; (4) \textsc{HypoEval}: pairwise arena ranking aggregated by Bradley--Terry--Davidson, complemented by rubric scoring as diagnostics.}
    \label{fig:pipeline}
\end{figure}

This section presents \textsc{HypoData}, the foundational data component of \textsc{HypoArena}, designed to systematically evaluate Prospective Hypothesis Discovery (PHD). We formalize hypothesis generation as a unified task across six domains, decoupling the model-visible \textbf{Context} from the hidden \textbf{Hypothesis} and its associated \textbf{Evidence}. Figure~\ref{fig:pipeline} illustrates the full data construction pipeline.
We formally define the task components and the mathematical formulation below.


\subsection{Task Definition}

\textit{Prospective Hypothesis Discovery} (PHD) is defined as the mapping from a reconstructed pre-conclusion context $\mathcal{C}$ to a \textit{Hypothesis Set} $\mathcal{H} = \{ (h_i, e_i) \}_{i=1}^{K}$. This task requires models to autonomously construct a plausible hypothesis space from inconclusive premises. Unlike traditional QA, which focuses on retrieving a pre-existing answer, PHD requires models to identify unresolved tensions and propose verifiable directions worth investigating next.

\paragraph{Context ($\mathcal{C}$).}
The context $\mathcal{C}$ approximates the raw, heterogeneous information state prior to conclusion formation, comprising anomalous observations, fragmented facts, and incomplete records. To preserve task validity, $\mathcal{C}$ must satisfy two constraints: \textbf{(i)}~\textit{informational sufficiency}, meaning it provides enough factual substrate to support non-trivial hypothesis generation; and \textbf{(ii)}~\textit{operational conclusion absence}, meaning that explicit final conclusions, target hypotheses, and retrospective causal attributions are withheld from the model-visible input. We refer to contexts satisfying the second criterion as operationally ``conclusion-free.''

\paragraph{Hypothesis Set ($\mathcal{H}$).}The output $\mathcal{H}$ consists of $(h_i, e_i)$ pairs, where each pair represents an atomic unit of reasoning:\begin{itemize}[leftmargin=1.5em, nosep]\item \textbf{Hypothesis ($h_i$)}: An explanatory or predictive claim that addresses the anomalies within $\mathcal{C}$. Each $h_i$ must be \textit{grounded} in the provided facts and \textit{verifiable} through subsequent testing. When $K>1$, the hypotheses should additionally be \textit{mutually discriminative}.\item \textbf{Evidence ($e_i$)}: The domain-specific justification for the hypothesis. In scientific research, this manifests as a \textit{verification plan} (e.g., experimental design); in investigative analysis, it takes the form of a \textit{diagnostic package} (e.g., actionable checks and supporting evidence logs) that substantiates the claim and guides further exploration.\end{itemize}


The cardinality $K$ of $\mathcal{H}$ is domain-dependent. Scientific cases elicit a single primary conjecture, emphasizing depth and testability, whereas analytical cases allow an open-cardinality set, emphasizing breadth and differentiation. A single-conjecture case can still admit multiple valid outputs across systems, so the source-derived Reference is not treated as a unique answer. 

\subsection{Data Collection}


To provide broad domain coverage and grounding in expert-authored sources, we curate source documents from six domains across two categories: \textbf{Scientific Domains} (Biomedical Science, Machine Learning, Social Science) and \textbf{Analytical Domains} (Financial Analysis, IT Operations, Safety Investigation). Domain-specific temporal cutoffs are enforced to minimize data contamination.
Detailed collection protocols are provided in Appendix~\ref{app:data-collection-details}.

\subsection{Data Construction with the Forge--Audit Agent Loop}

To balance the trade-off between information leakage and context richness, our framework employs an iterative \textit{Forge--Audit} loop. In this pipeline, the \textit{Forge} agent generates candidate contexts, while the \textit{Audit} agent validates them against benchmark constraints, repeating the cycle until criteria are met or the budget is exhausted. Detailed construction protocols and prompts are provided in Appendix~\ref{app:construction-details} and Appendix~\ref{app:prompts}. The \textit{Forge agent} operates through two sequential components:

\paragraph{Context Forge.}
The \textit{Context Forge} constructs a model-visible pre-conclusion context $\mathcal{C}$ from source-side factual material and, when appropriate, externally retrieved background. External search is constrained by source-specific timestamp boundaries that exclude materials dated after the relevant publication, release, or incident cutoff. These controls reduce the risk of introducing post-cutoff material through external retrieval. We apply different strategies based on the domain:

\begin{itemize}[leftmargin=1.5em, nosep]

\item \textbf{Scientific Domain:} To avoid reliance on any single, conclusion-bearing document, we retrieve a corpus of literature predating the discovery and use \textit{Document Merging} to synthesize multi-faceted sources into a coherent factual substrate.

\item \textbf{Analytical Domain:} To preserve raw evidence while stripping analytical bias, we employ \textit{Structural De-conclusion}. This process retains granular facts (e.g., timestamps and measurements) while systematically excising expert verdicts and explicit causal determinations.
\end{itemize}

\paragraph{Hypothesis Forge.}
The \textit{Hypothesis Forge} derives a source-grounded reference-side hypothesis set from the completed document while expressing each candidate in a form supportable from the reconstructed context. In scientific domains, this captures a central research conjecture or proposal core; in analytical domains, it yields distinct explanatory or diagnostic claims grounded in the factual record.

\paragraph{Audit Agent.}
After each Forge round, the \textit{Audit Agent} verifies the candidate against three checks: \textit{Leakage Check} tests whether the context inadvertently reveals the held-out hypothesis, \textit{Faithfulness Check} tests whether the hypothesis remains supported by the source document, and \textit{Supportability Check} tests whether the evidence provides sufficient grounding. Failed checks are returned to the Forge agent as actionable feedback for the next round.

    

\subsection{Human Quality Audit}

\begin{wraptable}{r}{0.4\textwidth}
\centering
\vspace{-\baselineskip}
\caption{Human Quality Audit Results.}
\label{tab:quality-audit}
\small
\renewcommand{\arraystretch}{1.3} 
\begin{tabular}{lc}
\toprule
\textbf{Dimension} & \textbf{Pass Rate} \\
\midrule
(a) Informativeness & 95\% \\
(b) Openness        & 100\% \\
(c) Completeness    & 100\% \\
(d) Supportiveness  & 97\% \\
\midrule
\textbf{Overall}    & \textbf{92\%} \\
\bottomrule
\end{tabular}
\end{wraptable}
To validate the fidelity of our automatically constructed instances, we sampled 20 cases from each of the three domains: Biomedical Science, Financial Analysis, and Social Science. For each domain, we recruited 2--3 expert annotators who are either senior doctoral candidates or practicing professionals. Each (Context, Hypothesis, Evidence) triple was evaluated against four criteria: Informativeness and Openness for Context, Completeness for Hypothesis, and Supportiveness for Evidence (detailed in Table~\ref{tab:quality-full}). Our results in Table~\ref{tab:quality-audit} show that 92\% of the sampled cases passed all criteria, providing evidence for construction quality on the audited subset. The full annotation guidelines are available in Appendix~\ref{app:annotation}.




The final benchmark represents each case as a model-visible, reconstructed pre-conclusion \textbf{Context} and a reference-side \textbf{Hypothesis Set ($\mathcal{H}$)} comprising hypothesis--evidence pairs $(h_i, e_i)$. The reference side is withheld during generation and is used for verification and diagnostic analysis or, in the arena, as an anonymous competitor for calibration. This separation supports evaluating a model's ability to synthesize disparate facts into grounded and verifiable hypotheses without exposing the source-derived Reference during generation.

\section{\textsc{HypoEval}: The Evaluation Framework}
\label{sec:hypoeval}
\textsc{HypoEval} evaluates submitted hypothesis sets against the shared model-visible context. The source-derived Reference is withheld from generation and enters the arena only as an anonymous competitor for calibration, not as a unique gold answer. This separation accommodates the open-ended nature of PHD, where a grounded hypothesis may remain valid even if it is absent from the Reference. Accordingly, \textsc{HypoEval} combines a primary \textbf{Rubric-based Arena} for relative ranking with \textbf{Rubric-based Scoring} for diagnostic analysis.

\subsection{Evaluation Rubrics}

Both tracks of \textsc{HypoEval} share a unified set of six metrics, with detailed definitions in Appendix~\ref{app:rubrics}, designed to assess hypothesis set quality across two levels.

\begin{itemize}[leftmargin=1.5em, nosep]
\item \textbf{Pair-level Fidelity}: For each $(h_i, e_i)$ pair, we score \textit{Contextual Grounding}, \textit{Inferential Insight}, and \textit{Evidential Justification}.
\item \textbf{Set-level Quality}: For submissions in domains that admit multiple pairs, we additionally score \textit{Hypothesis-Space Breadth}, \textit{Directional Distinctness}, and \textit{Analytical Utility}; \textit{Directional Distinctness} is not applicable when $K=1$.
\end{itemize}

\subsection{Rubric-based Arena (Primary Metric)}
\label{subsec:arena}

To overcome the ``reference-matching'' bottleneck and the biases of absolute scoring such as score compression and ceiling effects, we adopt a pairwise comparison protocol. In each matchup, an LLM-as-a-judge is presented with a shared conclusion-free context and the hypothesis sets generated by two models, A and B.
\begin{itemize}[leftmargin=1.5em, nosep]
\item \textbf{Position Debiasing.}
To mitigate position bias, each matchup is judged twice with A/B assignments swapped. The two verdicts are averaged into a single debiased score, and only matchups where both directions agree in polarity are marked as consistent.
\item \textbf{Aggregating Pairwise Judgments.}
The judge provides a 5-level categorical verdict: $A \gg B$, $A > B$, $A \approx B$, $B > A$, $B \gg A$, where $\gg$ indicates ``significantly better'' and $\approx$ indicates a tie. These verdicts are mapped to win-shares $w \in \{1.0, 0.75, 0.5, 0.25, 0.0\}$, respectively. To derive a global leaderboard, we aggregate outcomes using the Bradley--Terry--Davidson (BTD) model~\citep{davidson1970extending}, which models ties as a distinct epistemic state via a tie parameter $\theta$.

\end{itemize}

\subsection{Rubric-based Scoring (Secondary Metric)}

We complement the arena protocol with Rubric-based Scoring for diagnostic interpretability. In this track, the judge assigns independent absolute scores on a 1--5 scale to a model's output across the same six dimensions. We aggregate these into pair-level $Q_{\text{pair}}$ and set-level $Q_{\text{set}}$ scores, with full formulas in Appendix~\ref{app:rubrics}. This granular evaluation serves two purposes:

\begin{itemize}[leftmargin=1.5em, nosep]
\item \textbf{Multi-dimensional Profiling}: It produces per-dimension profiles that identify models strong in \textit{Contextual Grounding} but weak in \textit{Inferential Insight}, informing iterative model development.
\item \textbf{Complementary Absolute-Scale Context.} It provides an absolute-scale view of output quality, helping distinguish meaningful performance differences from relative advantages within an otherwise uniformly strong or weak model pool.
\end{itemize}

By synthesizing the discriminative power of competitive ranking with the diagnostic granularity of absolute scoring, \textsc{HypoEval} provides an assessment framework for evaluating discovery-oriented LLMs.

\section{Experiments}
\label{sec:experiments}




\begin{table}[t]
\centering
\caption{Main leaderboard under reference-independent pairwise judging by \texttt{seed-2.0-pro}. BTD initializes each model at a baseline rating of 1500 and updates ratings based on pairwise match outcomes. WinRate (WR) denotes each model's win fraction across all pairwise matchups. The Effort column reports each model's thinking-effort setting. Reference participates as an anonymous competitor for calibration.}
\label{tab:arena-seed}
\resizebox{\columnwidth}{!}{ 
\setlength{\tabcolsep}{3.2pt}
\begin{tabular}{cll ccc ccc cc}
\toprule
 & & & \multicolumn{3}{c}{\textbf{Scientific Domains}} & \multicolumn{3}{c}{\textbf{Analytical Domains}} & & \\
\cmidrule(lr){4-6} \cmidrule(lr){7-9}
\textbf{\#} & \textbf{Model} & \textbf{Effort} & \textbf{Bio} & \textbf{ML} & \textbf{Social} & \textbf{Fin} & \textbf{IT} & \textbf{Safety} & \textbf{Avg} & \textbf{WR} \\
\midrule
1 & claude-sonnet-4.6 & high & 1601.2 & 1579.3 & 1582.7 & \textbf{1739.2} & \textbf{1699.8} & \textbf{1721.6} & \textbf{1654.0} & 71.3\% \\
2 & claude-opus-4.6 & high & 1608.9 & 1618.9 & \textbf{1671.5} & 1670.3 & 1678.3 & 1668.4 & 1652.7 & 71.4\% \\
3 & gpt-5.4 & high & \textbf{1627.2} & \textbf{1650.6} & 1602.3 & 1636.2 & 1641.2 & 1630.7 & 1631.4 & 70.7\% \\
\rowcolor{colorGT} 4 & reference &  & 1568.8 & 1561.7 & 1502.2 & 1631.1 & 1604.8 & 1674.3 & 1590.5 & 57.8\% \\
5 & kimi-k2.6 & \checkmark & 1570.5 & 1562.4 & 1559.6 & 1601.0 & 1569.1 & 1633.5 & 1582.7 & 58.2\% \\
6 & glm-5.1 & \checkmark & 1592.8 & 1595.0 & 1591.6 & 1563.7 & 1588.3 & 1560.0 & 1581.9 & 59.1\% \\
7 & deepseek-v4-pro & high & 1525.0 & 1520.1 & 1564.8 & 1533.2 & 1517.5 & 1551.4 & 1535.4 & 46.1\% \\
8 & qwen-3.6-max & \checkmark & 1499.1 & 1518.6 & 1503.6 & 1513.2 & 1499.6 & 1520.0 & 1509.0 & 42.2\% \\
9 & minimax-m2.7 & \checkmark & 1508.3 & 1505.2 & 1492.5 & 1517.2 & 1516.1 & 1453.9 & 1498.9 & 37.0\% \\
10 & deepseek-v4-flash & high & 1464.3 & 1522.1 & 1513.8 & 1493.8 & 1424.6 & 1489.3 & 1484.6 & 36.3\% \\
11 & glm-5 & \checkmark & 1469.1 & 1450.2 & 1452.6 & 1485.0 & 1483.5 & 1445.2 & 1464.3 & 32.8\% \\
12 & minimax-m2.5 & \checkmark & 1458.4 & 1428.9 & 1427.4 & 1424.8 & 1433.4 & 1369.7 & 1423.8 & 22.9\% \\
13 & gpt-5.4-mini & high & 1424.0 & 1437.4 & 1393.7 & 1440.7 & 1397.7 & 1437.7 & 1421.9 & 23.7\% \\
14 & gemini-3.1-pro & high & 1429.9 & 1407.5 & 1415.7 & 1335.7 & 1388.2 & 1387.2 & 1394.0 & 20.7\% \\
15 & gemini-3-flash & high & 1420.0 & 1422.2 & 1418.2 & 1250.7 & 1301.0 & 1262.1 & 1345.7 & 14.7\% \\
16 & kimi-k2.5 & \checkmark & 1281.8 & 1246.5 & 1322.7 & 1266.1 & 1330.3 & 1289.4 & 1289.5 & 9.2\% \\
\bottomrule
\end{tabular}
} 
\end{table}

This section evaluates \textsc{HypoArena} along three axes: (i) the baseline leaderboard and the conditional effect of structured analytic skills (\S\ref{subsec:main_results}); (ii) arena vs.\ rubric scoring (\S\ref{subsec:arena-vs-score}); (iii) alignment with human annotation preference and external quality (\S\ref{subsec:alignment}).


\subsection{Experimental Setup}

We evaluate 15 contemporary LLMs on all 988 \textsc{HypoData} cases under two generation modes. These models span frontier closed-source and open-weight families, including \texttt{claude-sonnet-4.6}, \texttt{claude-opus-4.6}, and \texttt{gpt-5.4}. The full model list appears in Table~\ref{tab:arena-seed}. 

The primary evaluation follows the arena protocol (\S\ref{subsec:arena}) using \texttt{seed-2.0-pro} as judge. The source-derived Reference participates as an anonymous competitor for calibration. Cross-judge consistency with \texttt{mimo-v2-pro} is verified in Appendix~\ref{app:judge-triangulation}.



\begin{itemize}[leftmargin=1.5em, nosep]
\item \textbf{Baseline Mode}: The model generates its final Hypothesis Set directly from the input context in a single pass without invoking any explicit analytic skills. This mode serves as a zero-shot control, producing a concise single hypothesis--evidence pair for scientific domains, whereas for analytical domains, it yields an open-cardinality set of plausible hypotheses, serving as a performance floor for comparison.


\item \textbf{Agent Mode}: To mimic professional inquiry, models in this mode can select and execute a sequence from a library of 12 structured analytic skills (see Appendix~\ref{app:skills} for the full library). These skills are adapted from the methodology of \textit{Structured Analytic Techniques for Intelligence Analysis}~\citep{pherson2019structured}, including techniques such as the Analysis of Competing Hypotheses (ACH), Structured Brainstorming, and Chronology Analysis. Selected skills are composed into a sequential pipeline, where intermediate analytical artifacts are passed forward to refine the final submission.
\end{itemize}

\subsection{Main Results}
\label{subsec:main_results}

\paragraph{Clear performance stratification.}

\begin{figure}[t]
    \centering
    \vspace{-4pt}
    \includegraphics[width=0.95\columnwidth]{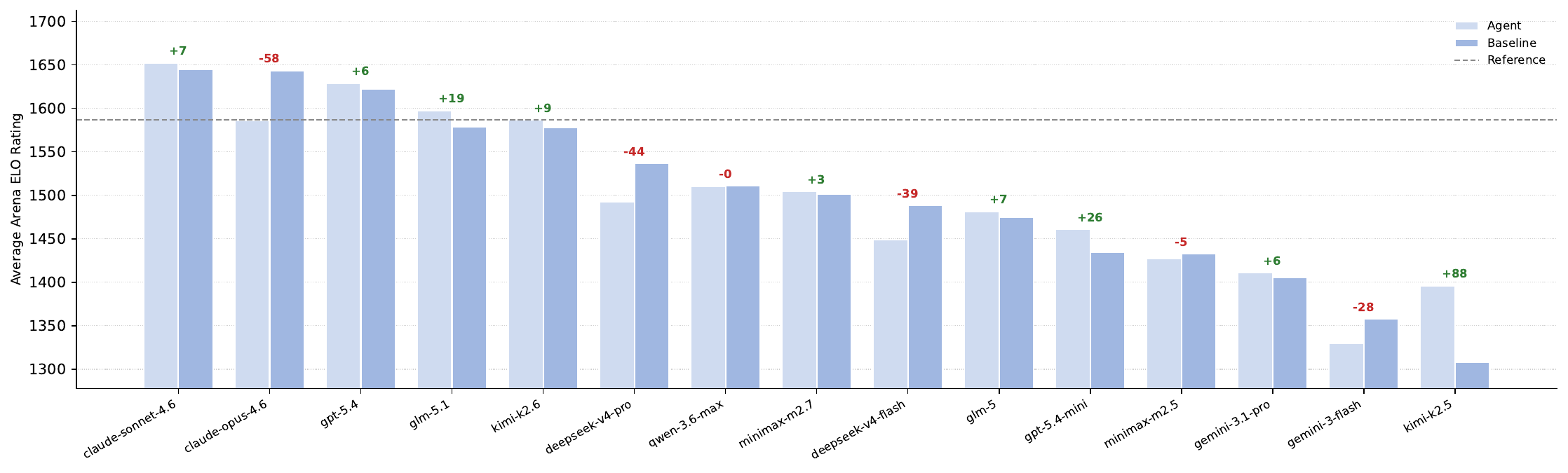}
    \caption{Per-model Agent vs.\ Baseline ratings on the same full-pool scale, sorted by baseline strength. Annotations show $\Delta = \text{Agent} - \text{Baseline}$ (green positive, red negative).}
    \vspace{-4pt}
    \label{fig:baseline-vs-agent}
    
\end{figure}

As shown in Table~\ref{tab:arena-seed}, the baseline leaderboard spans more than 360 BTD points, producing clear separation across the model pool. Three frontier models, \texttt{claude-sonnet-4.6}, \texttt{claude-opus-4.6}, and \texttt{gpt-5.4}, form a dominant top tier, leading across most domains. The protocol also resolves large within-family differences; for example, \texttt{kimi-k2.6} outscores \texttt{kimi-k2.5} by nearly 300 BTD points.

\paragraph{Reference as a domain-asymmetric baseline.}
The source-derived Reference outperforms most models under the primary judge but falls below the top frontier tier. Its strength is domain-dependent: it is more competitive in procedure-heavy domains such as Safety Investigation and Financial Analysis, but ranks lower in Social Science. This indicates that a fixed source-derived baseline is more competitive in domains with established procedural structure than in those admitting multiple analytical trajectories. 

\paragraph{Model-Dependent Effects of Structured Analytic Skills.}
As illustrated in Figure~\ref{fig:baseline-vs-agent}, Agent mode does not provide uniform gains. The shifts are widely dispersed, ranging from an 88-point gain for \texttt{kimi-k2.5} to a nearly 60-point decline for \texttt{claude-opus-4.6}, and show little monotonic association with baseline strength (Spearman's $\rho=-0.10$). These results indicate that structured analytic skills act as a model-dependent intervention, with candidate compression as one observed failure mode (Appendix~\ref{app:diagnostics}).


\subsection{Alignment with Human and External Signals}
\label{subsec:alignment}

\paragraph{Human Alignment and Judge Robustness.}
\begin{wrapfigure}[18]{r}{0.38\textwidth}
    \centering
    \includegraphics[width=0.98\linewidth]{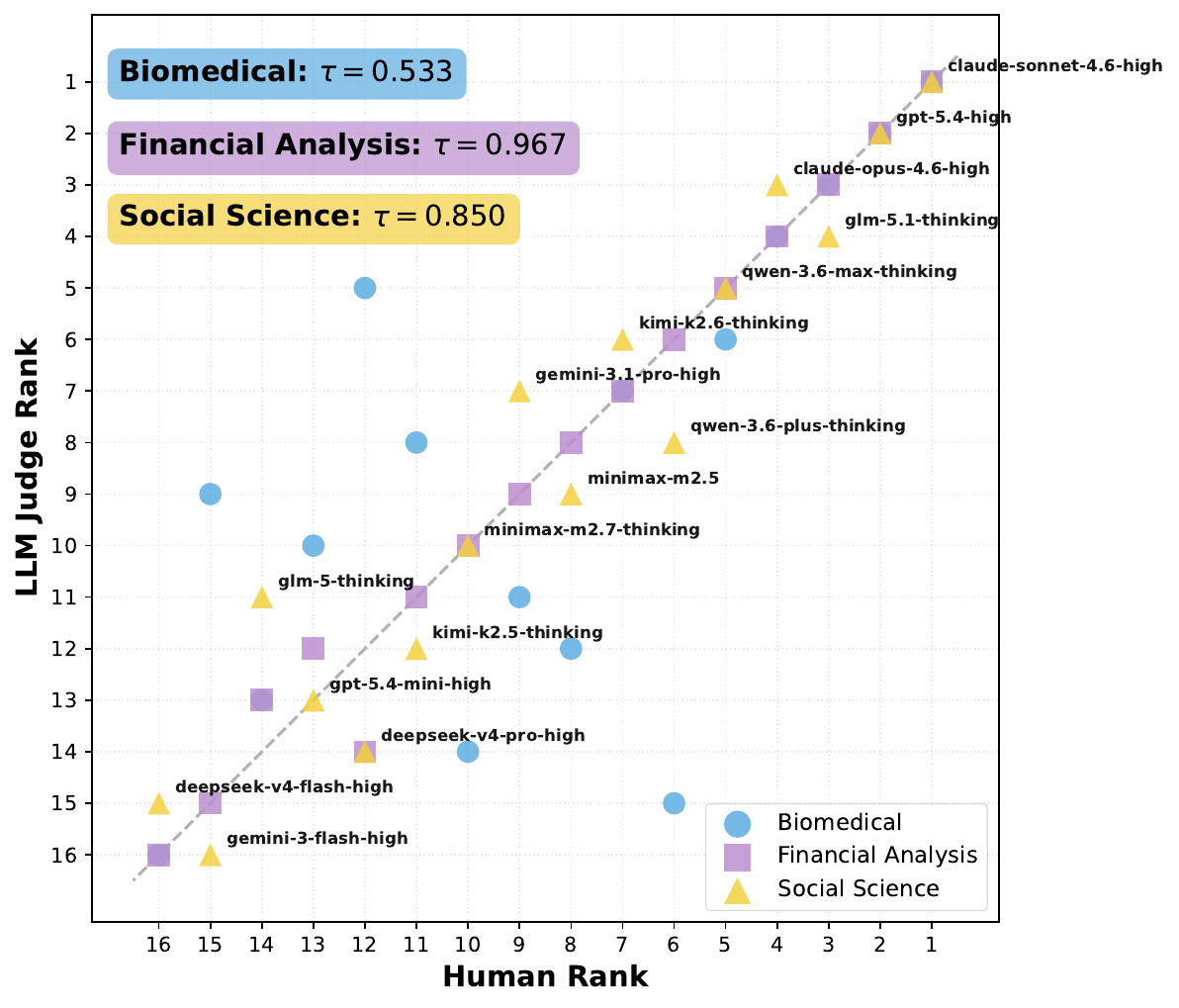}
    \caption{\textbf{Human vs. LLM Ranking Alignment.} Expert and primary-judge rankings show high overlap (cf.\ Table~\ref{tab:arena-per-domain}).}

    \label{fig:arena-comparison}
   
\end{wrapfigure}

To validate \textsc{HypoEval}, domain experts and our primary judge independently evaluated 1,500 pairwise comparisons. As visualized in Figure~\ref{fig:arena-comparison}, the resulting rankings exhibit strong global alignment (Kendall's $\tau=0.90$, Spearman's $\rho=0.98$). This consistency confirms that our automated arena protocol serves as a high-fidelity proxy for expert judgment. While we observe some per-domain variance, ranging from $\tau=0.97$ in Financial Analysis to $\tau=0.53$ in Biomedical Science due to interdisciplinary noise, the top-tier hierarchy remains identical across both evaluators. See Appendix~\ref{app:arena-judge-validation} for the full leaderboard.




\paragraph{Association with Peer-Review Outcomes.}
The Machine Learning subset was sampled from ICLR 2026 and stratified by acceptance outcome, an external signal not used by the arena judge. As detailed in Appendix~\ref{subapp:alignment-ml}, the source-derived Reference is more competitive on accepted than rejected cases across all three reported measures. Its bucket-specific BTD rating is 47 points higher, while the median per-case debiased score and win rate are higher by 0.06 and 0.04, respectively. This consistent directional pattern provides a complementary external check without treating acceptance outcome as ground truth for case-level hypothesis quality.
This convergence with an independent quality signal supports the external validity of our protocol.

\subsection{Pairwise Arena versus Direct Scoring}
\label{subsec:arena-vs-score}




\begin{figure}[t]
    \centering
    \includegraphics[width=0.95\columnwidth]{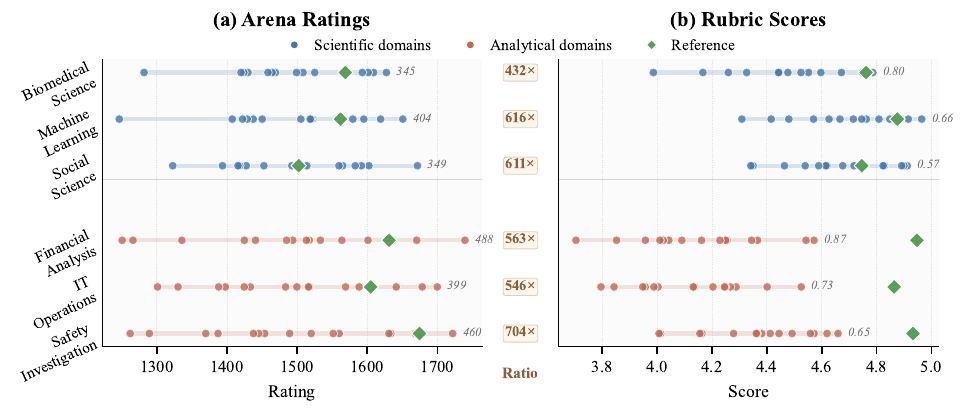}
    \caption{\textbf{Resolution contrast: Arena scaling vs. Rubric score compression.} While the arena protocol (left axis) provides high discriminative resolution across hundreds of points, traditional rubric scores (right axis) exhibit extreme \textbf{concentration}, with most models clustered within a \textbf{narrow 1-point band}.} 
    \label{fig:arena-vs-score}
\end{figure}

\paragraph{Score compression and ceiling effect.}
Both protocols agree on coarse ordering but diverge sharply at fine-grained resolution. As Figure~\ref{fig:arena-vs-score} shows, rubric scoring produces tighter model distributions: arena spreads range 345--490 points per domain, whereas rubric spreads stay under one point on the 1--5 scale, a compression ratio of 430$\times$--725$\times$. This compression is driven by a pronounced ceiling effect: in scientific domains, 95\%--98\% of assessments land at or above 4.0. Absent a single canonical answer, judges find it easier to declare one submission better in a head-to-head comparison than to commit to a low absolute score.

\paragraph{Protocol-dependent ranking differences.}
Score compression is accompanied by changes in fine-grained ordering, with per-domain Kendall $\tau$ between arena and rubric rankings ranging from 0.52 to 0.82. 
In the baseline pool, the Reference rises by three positions under rubric scoring in IT Operations, Financial Analysis, and Social Science, while \texttt{claude-opus-4.6} falls from its top-two arena placement in each of these domains. 
These shifts demonstrate sensitivity to the evaluation protocol, but rank changes alone do not identify the underlying mechanism.
Together with the score compression and ceiling effects described above, the protocol sensitivity motivates using arena comparisons for primary ranking while retaining rubric scores for diagnostics.

\section{Related Work}


\subsection{Benchmarks for Discovery-Oriented Tasks}
Recent benchmark-oriented work has begun to study scientific discovery and idea generation more systematically, moving beyond early generation systems toward explicit task formulation and standardized evaluation. Recent examples include inspiration-based scientific discovery benchmarks~\citep{research-bench}, principled benchmarking for hypothesis generation~\citep{hypo-bench}, research-idea generation benchmarks~\citep{ai-idea-bench}, and domain-specific fine-grained rediscovery settings~\citep{moose-chem-2}. Although these works differ in input format and scope, they collectively shift the field from demonstrating that LLMs can generate ideas or hypotheses to asking how such capabilities should be benchmarked, compared, and stress-tested. In parallel, another line of work has improved hypothesis-generation systems through literature-grounded or hybrid pipelines, while beginning to examine reliability issues such as grounding~\citep{literature-meets-data}, truthfulness~\citep{hyper}, and hallucination in generated scientific hypotheses~\citep{truth-hypo}. Relative to this literature, \textsc{HypoArena} centers on a particular combination of reconstructed pre-conclusion contexts under temporal and leakage controls, open-ended hypothesis generation without a unique reference answer, and unified coverage of scientific and analytical domains.

\subsection{Evaluation for Open-Ended Scientific Outputs}

Recent work has increasingly recognized that evaluating open-ended scientific outputs requires task-specific and knowledge-grounded protocols. Recent examples include arena-style evaluation platforms for non-verifiable scientific tasks~\citep{sci-arena}, multi-perspective frameworks for research-idea evaluation~\citep{inno-eval}, graph-based evaluators for idea assessment~\citep{graph-eval}, and work probing how grounded LLM critiques of scientific papers actually are~\citep{claim-check}. Collectively, these efforts move beyond treating scientific outputs as ordinary text-generation targets and instead emphasize evaluator reliability, grounding, and structured comparison. \textsc{HypoArena} addresses these by introducing \textit{Retrospective Context Regression} to filter explicit conclusion-bearing and retrospective causal content from source documents and \textsc{HypoEval} without relying on a single gold-standard reference.



\section{Conclusion}

We introduce \textsc{HypoArena}, a benchmark for \textit{Prospective Hypothesis Discovery (PHD)} that uses Retrospective Context Regression to reconstruct pre-conclusion discovery tasks from approximately 1,000 completed expert documents. Our \textsc{HypoEval} framework employs a pairwise arena and the Bradley--Terry--Davidson model to robustly evaluate open-ended hypothesis sets. Systematic testing of 15 frontier LLMs reveals heterogeneous, model-dependent effects of structured analytical skills, while qualitative generation traces identify candidate compression as a possible failure mode.

\bibliographystyle{unsrt}
\bibliography{styles/neurips_2026}

@misc{ai-idea-bench,
    title={AI Idea Bench 2025: AI Research Idea Generation Benchmark}, 
    author={Yansheng Qiu and Haoquan Zhang and Zhaopan Xu and Ming Li and Diping Song and Zheng Wang and Kaipeng Zhang},
    year={2025},
    eprint={2504.14191},
    archivePrefix={arXiv},
    primaryClass={cs.AI},
    url={https://arxiv.org/abs/2504.14191}, 
}

@inproceedings{claim-check,
    title = "{CLAIMCHECK}: How Grounded are {LLM} Critiques of Scientific Papers?",
    author = "Ou, Jiefu  and
      Walden, William  and
      Sanders, Kate  and
      Jiang, Zhengping  and
      Sun, Kaiser  and
      Cheng, Jeffrey  and
      Jurayj, William  and
      Wanner, Miriam  and
      Liang, Shaobo  and
      Morgan, Candice  and
      Han, Seunghoon  and
      Wang, Weiqi  and
      May, Chandler  and
      Recknor, Hannah  and
      Khashabi, Daniel  and
      Van Durme, Benjamin",
    editor = "Christodoulopoulos, Christos  and
      Chakraborty, Tanmoy  and
      Rose, Carolyn  and
      Peng, Violet",
    booktitle = "Findings of the Association for Computational Linguistics: EMNLP 2025",
    month = nov,
    year = "2025",
    address = "Suzhou, China",
    publisher = "Association for Computational Linguistics",
    url = "https://aclanthology.org/2025.findings-emnlp.1185/",
    doi = "10.18653/v1/2025.findings-emnlp.1185
        
        ",
    pages = "21712--21735",
    ISBN = "979-8-89176-335-7"
}

@inproceedings{discovery-bench,
    title={DiscoveryBench: Towards Data-Driven Discovery with Large Language Models},
    author={Bodhisattwa Prasad Majumder and Harshit Surana and Dhruv Agarwal and Bhavana Dalvi Mishra and Abhijeetsingh Meena and Aryan Prakhar and Tirth Vora and Tushar Khot and Ashish Sabharwal and Peter Clark},
    booktitle={The Thirteenth International Conference on Learning Representations},
    year={2025},
    url={https://openreview.net/forum?id=vyflgpwfJW}
}

@inproceedings{graph-eval,
    title={GraphEval: A Lightweight Graph-Based {LLM} Framework for Idea Evaluation},
    author={Tao Feng and Yihang Sun and Jiaxuan You},
    booktitle={The Thirteenth International Conference on Learning Representations},
    year={2025},
    url={https://openreview.net/forum?id=5RUM1aIdok}
}

@inproceedings{hyper,
    title = "{H}yp{ER}: Literature-grounded Hypothesis Generation and Distillation with Provenance",
    author = "Vasu, Rosni  and
      Basu, Chandrayee  and
      Dalvi Mishra, Bhavana  and
      Sarasua, Cristina  and
      Clark, Peter  and
      Bernstein, Abraham",
    editor = "Christodoulopoulos, Christos  and
      Chakraborty, Tanmoy  and
      Rose, Carolyn  and
      Peng, Violet",
    booktitle = "Proceedings of the 2025 Conference on Empirical Methods in Natural Language Processing",
    month = nov,
    year = "2025",
    address = "Suzhou, China",
    publisher = "Association for Computational Linguistics",
    url = "https://aclanthology.org/2025.emnlp-main.1292/",
    doi = "10.18653/v1/2025.emnlp-main.1292
        
        ",
    pages = "25413--25438",
    ISBN = "979-8-89176-332-6"
}

@misc{hypo-bench,
    title={HypoBench: Towards Systematic and Principled Benchmarking for Hypothesis Generation}, 
    author={Haokun Liu and Sicong Huang and Jingyu Hu and Yangqiaoyu Zhou and Chenhao Tan},
    year={2026},
    eprint={2504.11524},
    archivePrefix={arXiv},
    primaryClass={cs.AI},
    url={https://arxiv.org/abs/2504.11524}, 
}

@inproceedings{idea-bench,
    author = {Guo, Sikun and Shariatmadari, Amir Hassan and Xiong, Guangzhi and Huang, Albert and Kim, Myles and Williams, Corey M. and Bekiranov, Stefan and Zhang, Aidong},
    title = {IdeaBench: Benchmarking Large Language Models for Research Idea Generation},
    year = {2025},
    isbn = {9798400714542},
    publisher = {Association for Computing Machinery},
    address = {New York, NY, USA},
    url = {https://doi.org/10.1145/3711896.3737419},
    doi = {10.1145/3711896.3737419},
    booktitle = {Proceedings of the 31st ACM SIGKDD Conference on Knowledge Discovery and Data Mining V.2},
    pages = {5888–5899},
    numpages = {12},
    location = {Toronto ON, Canada},
    series = {KDD '25}
}

@misc{inno-eval,
    title={InnoEval: On Research Idea Evaluation as a Knowledge-Grounded, Multi-Perspective Reasoning Problem}, 
    author={Shuofei Qiao and Yunxiang Wei and Xuehai Wang and Bin Wu and Boyang Xue and Ningyu Zhang and Hossein A. Rahmani and Yanshan Wang and Qiang Zhang and Keyan Ding and Jeff Z. Pan and Huajun Chen and Emine Yilmaz},
    year={2026},
    eprint={2602.14367},
    archivePrefix={arXiv},
    primaryClass={cs.CL},
    url={https://arxiv.org/abs/2602.14367}, 
}

@inproceedings{insight-bench,
    title={InsightBench: Evaluating Business Analytics Agents Through Multi-Step Insight Generation},
    author={Gaurav Sahu and Abhay Puri and Juan A. Rodriguez and Amirhossein Abaskohi and Mohammad Chegini and Alexandre Drouin and Perouz Taslakian and Valentina Zantedeschi and Alexandre Lacoste and David Vazquez and Nicolas Chapados and Christopher Pal and Sai Rajeswar and Issam H. Laradji},
    booktitle={The Thirteenth International Conference on Learning Representations},
    year={2025},
    url={https://openreview.net/forum?id=ZGqd0cbBvm}
}

@inproceedings{literature-meets-data,
    title = "Literature Meets Data: A Synergistic Approach to Hypothesis Generation",
    author = "Liu, Haokun  and
      Zhou, Yangqiaoyu  and
      Li, Mingxuan  and
      Yuan, Chenfei  and
      Tan, Chenhao",
    editor = "Che, Wanxiang  and
      Nabende, Joyce  and
      Shutova, Ekaterina  and
      Pilehvar, Mohammad Taher",
    booktitle = "Proceedings of the 63rd Annual Meeting of the Association for Computational Linguistics (Volume 1: Long Papers)",
    month = jul,
    year = "2025",
    address = "Vienna, Austria",
    publisher = "Association for Computational Linguistics",
    url = "https://aclanthology.org/2025.acl-long.12/",
    doi = "10.18653/v1/2025.acl-long.12",
    pages = "245--281",
    ISBN = "979-8-89176-251-0"
}

@misc{moose-chem-2,
    title={MOOSE-Chem2: Exploring LLM Limits in Fine-Grained Scientific Hypothesis Discovery via Hierarchical Search}, 
    author={Zonglin Yang and Wanhao Liu and Ben Gao and Yujie Liu and Wei Li and Tong Xie and Lidong Bing and Wanli Ouyang and Erik Cambria and Dongzhan Zhou},
    year={2025},
    eprint={2505.19209},
    archivePrefix={arXiv},
    primaryClass={cs.CL},
    url={https://arxiv.org/abs/2505.19209}, 
}

@misc{research-bench,
    title={ResearchBench: Benchmarking LLMs in Scientific Discovery via Inspiration-Based Task Decomposition}, 
    author={Yujie Liu and Zonglin Yang and Tong Xie and Jinjie Ni and Ben Gao and Yuqiang Li and Shixiang Tang and Wanli Ouyang and Erik Cambria and Dongzhan Zhou},
    year={2025},
    eprint={2503.21248},
    archivePrefix={arXiv},
    primaryClass={cs.CL},
    url={https://arxiv.org/abs/2503.21248}, 
}

@inproceedings{sci-arena,
    title={SciArena: An Open Evaluation Platform for Non-Verifiable Scientific Literature-Grounded Tasks},
    author={Yilun Zhao and Kaiyan Zhang and Tiansheng Hu and Sihong Wu and Ronan Le Bras and Yixin Liu and Xiangru Tang and Joseph Chee Chang and Jesse Dodge and Jonathan Bragg and Chen Zhao and Hannaneh Hajishirzi and Doug Downey and Arman Cohan},
    booktitle={The Thirty-ninth Annual Conference on Neural Information Processing Systems Datasets and Benchmarks Track},
    year={2025},
    url={https://openreview.net/forum?id=am6RR85mnc}
}

@inproceedings{truth-hypo,
    title     = {Toward Reliable Scientific Hypothesis Generation: Evaluating Truthfulness and Hallucination in Large Language Models},
    author    = {Xiong, Guangzhi and Xie, Eric and Williams, Corey and Kim, Myles and Shariatmadari, Amir Hassan and Guo, Sikun and Bekiranov, Stefan and Zhang, Aidong},
    booktitle = {Proceedings of the Thirty-Fourth International Joint Conference on
               Artificial Intelligence, {IJCAI-25}},
    publisher = {International Joint Conferences on Artificial Intelligence Organization},
    editor    = {James Kwok},
    pages     = {7849--7857},
    year      = {2025},
    month     = {8},
    note      = {Main Track},
    doi       = {10.24963/ijcai.2025/873},
    url       = {https://doi.org/10.24963/ijcai.2025/873},
}

@article{wang2026fire,
  title={FIRE-Bench: Evaluating Agents on the Rediscovery of Scientific Insights},
  author={Wang, Zhen and Bai, Fan and Luo, Zhongyan and Su, Jinyan and Sun, Kaiser and Yu, Xinle and Liu, Jieyuan and Zhou, Kun and Cardie, Claire and Dredze, Mark and others},
  journal={arXiv preprint arXiv:2602.02905},
  year={2026}
}

@article{egg2025dabstep,
  title={Dabstep: Data agent benchmark for multi-step reasoning},
  author={Egg, Alex and Goyanes, Martin Iglesias and Kingma, Friso and Mora, Andreu and von Werra, Leandro and Wolf, Thomas},
  journal={arXiv preprint arXiv:2506.23719},
  year={2025}
}

@article{jingdsbench,
  title={DSBench: How Far Are Data Science Agents from Becoming Data Science Experts?},
  author={Jing, Liqiang and Huang, Zhehui and Wang, Xiaoyang and Yao, Wenlin and Yu, Wenhao and Ma, Kaixin and Zhang, Hongming and Du, Xinya and Yu, Dong},
  journal={arXiv preprint arXiv:2409.07703},
  year={2024}
}

@article{sican,
  title={Can llms generate novel research ideas? a large-scale human study with 100+ nlp researchers},
  author={Si, Chenglei and Yang, Diyi and Hashimoto, Tatsunori},
  journal={arXiv preprint arXiv:2409.04109},
  year={2024}
}

@inproceedings{lin2004rouge,
  title={Rouge: A package for automatic evaluation of summaries},
  author={Lin, Chin-Yew},
  booktitle={Text summarization branches out},
  pages={74--81},
  year={2004}
}

@inproceedings{papineni2002bleu,
  title={Bleu: a method for automatic evaluation of machine translation},
  author={Papineni, Kishore and Roukos, Salim and Ward, Todd and Zhu, Wei-Jing},
  booktitle={Proceedings of the 40th annual meeting of the Association for Computational Linguistics},
  pages={311--318},
  year={2002}
}

@inproceedings{liu2023g,
  title={G-eval: NLG evaluation using gpt-4 with better human alignment},
  author={Liu, Yang and Iter, Dan and Xu, Yichong and Wang, Shuohang and Xu, Ruochen and Zhu, Chenguang},
  booktitle={Proceedings of the 2023 conference on empirical methods in natural language processing},
  pages={2511--2522},
  year={2023}
}

@article{mitchener2025kosmos,
  title={Kosmos: An ai scientist for autonomous discovery},
  author={Mitchener, Ludovico and Yiu, Angela and Chang, Benjamin and Bourdenx, Mathieu and Nadolski, Tyler and Sulovari, Arvis and Landsness, Eric C and Barabasi, Daniel L and Narayanan, Siddharth and Evans, Nicky and others},
  journal={arXiv preprint arXiv:2511.02824},
  year={2025}
}

@article{davidson1970extending,
  title={On extending the Bradley-Terry model to accommodate ties in paired comparison experiments},
  author={Davidson, Roger R},
  journal={Journal of the American Statistical Association},
  volume={65},
  number={329},
  pages={317--328},
  year={1970},
  publisher={Taylor \& Francis}
}

@book{pherson2019structured,
  title={Structured analytic techniques for intelligence analysis},
  author={Pherson, Randolph H and Heuer Jr, Richards J},
  year={2019},
  publisher={Cq Press}
}

\clearpage
\beginappendix

\section{Limitations}
\label{app:limitations}

Despite its comprehensive framework, \textsc{HypoArena} has several limitations that open avenues for future research. First, while our benchmark spans six diverse domains, it remains primarily text-centric. Expanding \textsc{HypoArena} to include multimodal or structured-data-driven discovery settings, such as genomics and medical imaging, represents a critical next step. Second, the current construction pipeline relies on a single model (\texttt{gpt-5.4}) for the \textit{Forge} process. Future iterations could incorporate a more diverse ensemble of construction models or human-authored cases to enhance stylistic variety and mitigate potential model-specific biases. Third, although we verified evaluation consistency with a second judge, the reliability of our framework could be further strengthened by employing a broader judge panel or a hybrid human-LLM scoring system. Finally, the \textit{Agent Mode} currently operates within a fixed library of analytic skills. Moving toward a more flexible paradigm where models can dynamically compose or define novel strategies would better reflect the fluid nature of human scientific inquiry and investigative reasoning.
\section{Details of Data Collection}
\label{app:data-collection-details}

To provide broad domain coverage and grounding in expert-authored sources, \textsc{HypoData} spans six domains organized into two high-level categories: \textit{Scientific Domains} (Biomedical Science, Machine Learning, and Social Science) and \textit{Analytical Domains} (Financial Analysis, IT Operations, and Safety Investigation). Across all domains, we prioritize publicly accessible and authoritative source materials that are rich enough to support context reconstruction and source-grounded construction of reference-side hypothesis--evidence pairs. Where temporal control is important, we further restrict collection to recent artifacts or apply domain-appropriate release-date constraints to reduce contamination while preserving sufficient source diversity. Unless otherwise noted, each collected source artifact is processed into a single final benchmark case, whose reference side may contain one or more hypothesis--evidence pairs.

\paragraph{Biomedical Science.}
Biomedical Science sources are primary research articles retrieved from PubMed\footnote{\url{https://pubmed.ncbi.nlm.nih.gov}} via a systematic query pipeline, restricted to open-access full texts published between July 2025 and April 2026 from high-impact journals including \textit{Nature Communications}, \textit{Cell Reports}, \textit{eLife}, \textit{Nucleic Acids Research}, \textit{PLOS Biology}, \textit{PNAS}, and \textit{Science Advances}. We retain only empirical papers with explicit hypothesis framing, excluding reviews, case reports, clinical trials, and letters, yielding 244 benchmark cases.

\paragraph{Machine Learning.}
Machine Learning draws on ICLR 2026\footnote{\url{https://openreview.net/group?id=ICLR.cc/2026}} submissions with released final decisions. To diversify research maturity and benchmark difficulty, we stratify by decision outcomes, yielding 218 cases. Decision categories shape sampling only and are not used as supervision targets.

\paragraph{Social Science.}

Social Science comprises 163 articles from seven journals: \textit{Journal of Marketing Research}\footnote{\url{https://onlinelibrary.wiley.com/journal/15206793}}, \textit{Journal of Retailing and Consumer Services}\footnote{\url{https://www.sciencedirect.com/journal/journal-of-retailing-and-consumer-services}}, \textit{Journal of International Business Studies}\footnote{\url{https://link.springer.com/journal/41267}}, \textit{Journal of Business Research}\footnote{\url{https://www.sciencedirect.com/journal/journal-of-business-research}}, \textit{Journal of Consumer Research}\footnote{\url{https://onlinelibrary.wiley.com/journal/1745459x}}, \textit{Journal of Consumer Psychology}\footnote{\url{https://myscp.onlinelibrary.wiley.com/journal/15327663}}, and \textit{Annals of Tourism Research}\footnote{\url{https://www.sciencedirect.com/journal/annals-of-tourism-research}}, restricted to 2025--2026 publications. Each paper is verified by domain experts to contain explicit hypothesis-bearing argumentation suitable for context reconstruction.

\paragraph{Financial Analysis.}
Financial Analysis instances are anchored on the quarterly disclosure cycle of large-cap U.S.-listed companies, combining primary 10-Q filings (2025--2026) sourced from the SEC\footnote{\url{https://www.sec.gov/search-filings}} with matched publicly accessible professional analyses. Context is reconstructed primarily from the filing, while paired analyses supply hypothesis-level interpretation and supporting evidence. Cases are retained only when the quarterly disclosure is substantive and the analytical coverage is deep, yielding 114 cases.

\paragraph{IT Operations.}
IT Operations sources are public post-mortem reports from widely referenced incident repositories\footnote{\url{https://github.com/danluu/post-mortems}} and engineering blogs of major technology companies. Unlike other domains, no strict temporal cutoff is imposed: post-mortem reports describe unique incident configurations and system-specific failure chains that are unlikely to appear verbatim in model training data, making temporal contamination a lower risk than in publication-based domains. We retain technically detailed cases with explicit descriptions of failure mechanisms, contributing factors, or remediation-relevant reasoning, yielding 146 cases.

\paragraph{Safety Investigation.}
Safety Investigation cases are drawn from official U.S. government accident investigation reports, including transportation accident reports\footnote{\url{https://www.ntsb.gov/investigations/AccidentReports/Pages/Reports.aspx}} and industrial chemical incident investigations\footnote{\url{https://www.csb.gov/news/incident-report-rule-form-}}, restricted to reports released in 2025 or later. We retain only cases with detailed event narratives and causal analyses to support mechanistic reasoning, yielding 103 cases.

\begin{table}[t]
    \centering
    \small
    \setlength{\tabcolsep}{2pt}
    \renewcommand{\arraystretch}{1}
    \caption{Statistics and source overview of \textsc{HypoData}. Each hypothesis--evidence (H-E) pair is the atomic evaluation unit. Scientific domains yield focused, single-direction hypotheses; real-world domains yield multi-directional hypothesis sets. Detailed collection protocols are in Appendix~\ref{app:data-collection-details}.}
    \label{tab:dataset-statistics}
    \begin{tabular}{llccc}
        \toprule
        \textbf{Domain} & \textbf{Primary Source} & \textbf{\# Cases} & \textbf{\# H-E} & \textbf{\# Category} \\
        \midrule
\textit{Research} & & & & \\
\quad Biomedical Science & Journal Articles & 244 & 244 &  \\
\quad Machine Learning & Conference Papers & 218 & 218 &  \\
\quad Social Science & Journal Articles & 163 & 163 &  \\
\midrule
\textit{Real-World} & & & & \\
\quad Financial Analysis & SEC 10-Q + Analyst Reports & 114 & 456 & 3.99 \\
\quad IT Operations & Post-mortems Blogs & 146 & 527 & 3.29 \\
\quad Safety Investigation & CSB \& NTSB Reports & 103 & 404 & 3.87 \\
\midrule
\textbf{Total} & & \textbf{988} & \textbf{2{,}012} & \textbf{3.67} \\
\bottomrule
    \end{tabular}
\end{table}

\section{Details of Data Construction}
\label{app:construction-details}

This appendix provides the full implementation details of the two-stage Forge--Audit pipeline used to construct \textsc{HypoData} (Section~\ref{sec:hypodata}).

\subsection{Forge--Audit Pipeline}

Each benchmark instance is constructed through an iterative multi-agent pipeline. A coordinating controller dispatches specialized Forge and Audit agents across both construction stages. The Forge agent generates candidate outputs (contexts in Stage~1; hypothesis--evidence pairs in Stage~2), while the Audit agent evaluates whether the candidates satisfy benchmark requirements. If the candidate passes audit, construction terminates; otherwise, the Audit agent returns explicit and actionable feedback, which is used as direct input to the next Forge round. Construction proceeds through repeated cycles until the audit passes or a fixed iteration budget is reached. Under this protocol, each collected source artifact is processed exactly once and yields a single final benchmark instance.

\subsection{Stage 1: Context Construction under Temporal Control}

\paragraph{Domain-Specific Construction Policies.}
Context construction follows different default policies across domains. In scientific domains, where source papers often provide limited background and related work, the Forge agent may enrich the context through temporally controlled search over academic and web sources. In real-world domains, where source materials are typically more information-dense, context is constructed primarily from the source materials themselves, with minimal or no external augmentation. In these domains, the construction process places greater emphasis on preserving source-side factual detail and, when possible, the original writing texture.

\paragraph{Strict Temporal Cutoff.}
When external retrieval is used, all search tools are constrained by source-specific timestamp boundaries so that only information available prior to the publication, release, or incident date of the source materials can be accessed. This reduces leakage from future knowledge and helps preserve the temporal realism of the original reasoning setting.

\paragraph{Dynamic Context Reconstruction.}
Directly extracting localized arguments from the source materials often yields overly prescriptive inputs and leads to homogenized hypothesis generation. To preserve exploratory reasoning, the Forge agent reconstructs context by combining source-side factual content with temporally admissible external background when appropriate. During this process, it removes explicit hypothesis statements, conclusion-side claims, citation anchors, and other cues that would reveal the original line of reasoning. The resulting context is designed to be information-rich, multidimensional, coherent, and pre-conclusion in character, while avoiding answer-side guidance.

\subsection{Stage 2: Hypothesis and Evidence Extraction}

While the main text characterizes the \textit{target} of Stage~2 along two patterns---vertical deepening and horizontal association---the \textit{extraction method} varies by source structure.

\paragraph{Hypothesis Extraction Strategies.}
Because hypothesis presentation varies substantially across domains, the Forge agent applies one of three strategies based on the full source materials:
\begin{itemize}
    \item \textbf{Direct Extraction:} When the source explicitly states a single central hypothesis, it is extracted with minimal modification.
    \item \textbf{Synthesis and Rewriting:} When multiple distributed sub-hypotheses appear across the source, they are consolidated into a unified core claim.
    \item \textbf{Latent Distillation:} When no explicit hypothesis is stated, the underlying claim is inferred from the source's methods, results, and conclusions.
\end{itemize}

\paragraph{Evidence Construction.}
Each evidence field $e_i$ provides the domain-specific support associated with hypothesis $h_i$. As an umbrella field, it may take either prospective form (verification sketches in scientific domains) or retrospective form (evidence packages in real-world domains). By constructing this field explicitly, we require benchmark targets to be not only plausible but also supportable.

\subsection{Input--Reference Boundary Enforcement}

Throughout construction, we strictly separate the model-input side from the reference side of each benchmark case. The context must not leak final conclusions, posterior explanations, or direct restatements of the source's claims; all hypothesis--evidence pairs remain exclusively on the reference side. This boundary is enforced explicitly in the Audit loop and is central to maintaining the benchmark's pre-conclusion setting.



\section{Human Annotation Details}
\label{app:annotation}

To assess the construction quality of \textsc{HypoData} and the alignment of our automated evaluator with expert preferences, we designed two distinct annotation tasks, \textbf{Quality Audit} and \textbf{Preference Evaluation}. All tasks were performed in English by domain experts. This section describes the annotation guidelines and setup.

\subsection{General Experimental Setup}

The annotation setup was as follows.

\begin{itemize}[leftmargin=1.5em, itemsep=2pt]

\item \textbf{Expertise}. We recruited 2--3 annotators per domain, consisting of senior Ph.D. candidates or practicing professionals in Biomedical Science, Financial Analysis, and Social Science.
\item \textbf{Anonymization}. To reduce identity- and provenance-related bias, model identities and reference provenance were hidden from annotators where applicable.
\item \textbf{Data Scale}. Contexts averaged $\sim$3,000 tokens, while Hypothesis--Evidence pairs averaged $\sim$1,000 tokens.

\end{itemize}

\subsection{Quality Audit}
\paragraph{Objective.} Assess whether the automatically constructed (Context, Hypothesis, Evidence) triples provide informative and open-ended contexts, complete source-grounded hypotheses, and supportive evidence.
\paragraph{Input Format.} Annotators are presented with a Context and its corresponding (Hypothesis, Evidence) tuples. In scientific domains, each case contains a single tuple; in real-world analytical domains, multiple tuples may be provided.
\paragraph{Evaluation Criteria.} Annotators perform four binary checks across three core dimensions (detailed in Table~\ref{tab:quality-full}).

A key qualitative metric is Context Openness: This measures whether the Context allows for multiple plausible analytical directions. While a "No" (indicating a singular reasonable direction) is acceptable for specific deterministic cases, a high "Yes" rate across the dataset is preferred to ensure the benchmark's challenge level.

\subsection{Preference Evaluation}
\label{app:preference-evaluation}
\paragraph{Objective.} Collect pairwise human preference judgments as an expert reference signal for assessing the alignment of our automated LLM-as-a-judge system.
\paragraph{Input Format.} For each instance, annotators are provided with:

\begin{enumerate}[nosep]
\item \textbf{Shared Context}. The conclusion-free background information provided to both models.
\item \textbf{Model Outputs}. Two candidate sets of hypotheses and evidence (Output A and Output B), randomized and stripped of model metadata.
\item \textbf{Evaluation Metrics}. The standardized rubric used for assessment (see Appendix~\ref{app:rubrics}).
\end{enumerate}

\paragraph{Labeling Requirements.} For each pairwise matchup, annotators must provide:
\begin{itemize} 
\item \textbf{Preference}. Select from {A $>$ B, Tie, B $>$ A}.
\item \textbf{Confidence Score}. Annotators rated their confidence as High, Medium, or Low.
\end{itemize}

\subsection{Results of Human Annotation}
Results of the Quality Audit and Preference Evaluation are reported in Table~\ref{tab:quality-audit-results} and Appendix~\ref{app:arena-judge-validation}, respectively.

\begin{table}[ht]
\centering
\small
\caption{Summary of Quality Audit Criteria. These questions guide human annotators in assessing whether the synthesized Context, Hypothesis, and Evidence meet the intended quality standards.}
\begin{tabularx}{0.9\textwidth}{cllX}
\toprule
\textbf{ID} & \textbf{Criterion} & \textbf{Dim.} & \textbf{Core Audit Question} \\
\midrule
(a) & Informativeness & Context & Does the context provide sufficient background and key facts to guide hypothesis generation? \\
\addlinespace
(b) & Openness & Context & Can the context elicit diverse and plausible alternative hypotheses beyond the target? \\
\addlinespace
(c) & Completeness & Hypothesis & Does the hypothesis capture the core claims originally presented in the source document? \\
\addlinespace
(d) & Supportiveness & Evidence & Does the evidence provide a sound verification plan or factual basis for the hypothesis? \\
\bottomrule
\end{tabularx}

\label{tab:quality-full}
\end{table}

\begin{table}[ht]
\centering
\small 

\setlength{\tabcolsep}{4pt} 
\caption{Human Quality Audit Results. Criteria (a--d) correspond to Informativeness, Openness, Completeness, and Supportiveness respectively. The \textbf{92\%} overall pass rate validates the robustness of our HypoData construction.}
\begin{tabular}{lcccccc}
\toprule
\textbf{Domain} & \textbf{a} & \textbf{b} & \textbf{c} & \textbf{d} & \textbf{Pass Rate} \\
\midrule
Biomedical       & 85\%  & 100\% & 100\% & 95\%  & 80\% \\
Financial        & 100\% & 100\% & 100\% & 100\% & 100\% \\
Social Science   & 100\% & 100\% & 100\% & 95\%  & 95\% \\
\midrule
\textbf{Overall}  & \textbf{95\%} & \textbf{100\%} & \textbf{100\%} & \textbf{97\%} & \textbf{92\%} \\
\bottomrule
\end{tabular}

\label{tab:quality-audit-results}
\end{table}

\section{Generation-Side Diagnostics}
\label{app:diagnostics}

Qualitative inspection of the generation-side logs suggests three recurrent distinctions among systems that may help explain why models appearing similarly plausible on the surface can diverge substantially under the primary evaluation protocol.

\paragraph{Coverage breadth.}
In multi-pair domains, the strongest systems do not simply produce more pairs; they produce more meaningfully distinct pairs. Their advantage comes from broader exploration of the supported hypothesis space while maintaining non-redundancy and local quality. Weaker systems often generate only a small set of familiar directions, even when the context supports a wider range of mechanistic, latent, or intervention-oriented possibilities.

\paragraph{Candidate compression.}
In several inspected traces, particularly under Agent Mode, systems generate rich intermediate analyses but compress them into overly narrow final submissions. These systems identify multiple plausible lines of reasoning internally yet fail to preserve them in the submitted set. This possible failure mode is especially costly in open-ended investigative settings, where omitted candidates may correspond to important supported directions rather than merely stylistic alternatives.

\paragraph{Mechanism granularity.}
Strong systems formulate hypotheses at an appropriate level of explanatory resolution: they identify sharper mechanisms, separate distinct causal pathways, and attach concrete support to each direction. Weaker systems often conflate multiple mechanisms into a single diffuse hypothesis or remain at a level of generality too broad to be meaningfully testable. Such differences may be muted under direct scoring but become highly visible under arena comparison and set-level evaluation.

Taken together, these observations suggest that failure in deep hypothesis generation is not always reducible to obvious factual mistakes or low fluency. It may instead reflect insufficient breadth, insufficient decomposition, or loss of structure between intermediate reasoning and the final submitted set.

\section{Details of Evaluation}
\label{app:evaluation-details}

\subsection{Submission Cardinality}

\textsc{HypoEval} does not impose a hard cap on submission size in multi-pair domains. This reflects the fact that breadth of supported hypothesis exploration is itself part of the target capability. At the same time, larger submissions are not automatically preferred: additional pairs improve evaluation only insofar as they expand meaningful coverage without reducing distinctness or local quality. All formal evaluation is therefore conducted on the final submitted set.

\subsection{Details of the Rubric}
\label{app:rubrics}
\paragraph{Quality Dimensions.}
Evaluation operates at two granularities to handle both local plausibility and global coherence. At the \textit{pair level}, each submitted $(\hat{h}_i, \hat{e}_i)$ is assessed on three dimensions:
\begin{itemize}[leftmargin=1.5em, nosep]
\item \textit{Contextual Grounding}: whether the hypothesis is anchored in specific facts, observations, or unresolved tensions in $\mathcal{C}$, and whether the inference is warranted---not merely topically relevant.
\item \textit{Inferential Insight}: whether the hypothesis goes beyond surface restatement by synthesizing dispersed information into a non-obvious, context-constrained explanatory, predictive, or mechanistic claim.
\item \textit{Evidential Justification}: whether the evidence concretely and proportionately supports the hypothesis, with a coherent reasoning chain and without overclaiming.
\end{itemize}
These dimensions operationalize the requirement that hypotheses be grounded, non-trivial, and supportable (Section~\ref{sec:hypodata}). When $K>1$, the submission is additionally evaluated at the \textit{set level}:
\begin{itemize}[leftmargin=1.5em, nosep]
\item \textit{Hypothesis-Space Breadth}: coverage of multiple distinct, context-supported directions such as mechanisms, risk axes, or causal pathways.
\item \textit{Directional Distinctness}: non-redundancy across hypotheses---not overlapping or trivially reformulated versions of the same claim.
\item \textit{Analytical Utility}: actionability for downstream investigation, testing, or prioritization.
\end{itemize}
Set-level evaluation is necessary because a locally plausible but narrow or repetitive submission may still be materially incomplete.

\subsection{Rubric Scoring Formulas}

For each submitted pair $(\hat{h}_i,\hat{e}_i)$, the judge assigns scores on Contextual Grounding ($g_i$), Inferential Insight ($\ell_i$), and Evidential Justification ($j_i$). We define the local pair score as
\[
q_i=\frac{g_i+\ell_i+j_i}{3}.
\]
Pair-level quality is aggregated over all submitted pairs as
\[
Q_{\text{pair}}=\frac{1}{K}\sum_{i=1}^{K}q_i.
\]
This average reflects the overall quality of the submitted hypothesis--evidence pairs and prevents larger submissions from being rewarded solely for containing more candidates.

For domains that admit multi-pair submissions, the judge additionally assigns set-level scores on Hypothesis-Space Breadth ($b$), Directional Distinctness ($d$), and Analytical Utility ($u$). For submissions with $K>1$, we define
\[
Q_{\text{set}}=\frac{b+d+u}{3}.
\]
If a system submits only a single pair in a domain that admits multi-pair submissions, Directional Distinctness is not applicable, and we compute
\[
Q_{\text{set}}=\frac{b+u}{2}.
\]

In reporting rubric-based results, we primarily present $Q_{\text{pair}}$ and $Q_{\text{set}}$ separately. When a compact summary is useful, we report $S_{\text{rubric}}=q_1$ for singleton domains and $S_{\text{rubric}}=(Q_{\text{pair}}+Q_{\text{set}})/2$ for domains that admit multi-pair submissions. This summary is used only for concise diagnostic reporting and not for primary model ranking.

\section{Full-Pool Arena and Rubric Leaderboards}
\label{app:full-pool-leaderboards}

This appendix reports both protocols at the (model, mode) level on a single full-pool BTD scale, so that Section~\ref{subsec:arena-vs-score}'s cascade can be inspected row by row against specific numbers. Table~\ref{tab:arena-all-seed} ranks every (model, mode) cell by arena BTD; Table~\ref{tab:score-seed} reports the per-(model, mode) mean rubric score $S$ in the same row order, with a \textit{Rubric Rank} column showing how the same set is re-ranked under rubric Avg $S$; Table~\ref{tab:score-details-fin-seed} reports the detailed results of different Rubrics in Financial Analysis.

\begin{table}[t]
\centering
\caption{Full-pool arena leaderboard under \texttt{seed-2.0-pro}: every (model, mode) cell on the same BTD scale across the six domains. Rows are sorted by Avg BTD descending; \colorbox{colorGT}{reference} participates as a single anonymous competitor (no agent counterpart). This table is the arena-side counterpart to Table~\ref{tab:score-seed} and the source of the \textit{Arena \#} ordering used there.}
\label{tab:arena-all-seed}
\small
\setlength{\tabcolsep}{3.5pt}
\begin{tabular}{cll rrr rrr rr}
\toprule
 & & & \multicolumn{3}{c}{\textbf{Scientific Domains}} & \multicolumn{3}{c}{\textbf{Analytical Domains}} & & \\
\cmidrule(lr){4-6} \cmidrule(lr){7-9}
\textbf{\#} & \textbf{Model} & \textbf{Mode} & \textbf{Bio} & \textbf{ML} & \textbf{Social} & \textbf{Fin} & \textbf{IT} & \textbf{Safety} & \textbf{Avg} & \textbf{WR} \\
\midrule
1 & claude-sonnet-4.6 & Agent & 1580.5 & 1594.5 & 1612.2 & \textbf{1754.1} & 1672.9 & \textbf{1700.2} & \textbf{1652.4} & 73.6\% \\
2 & claude-sonnet-4.6 & Baseline & 1609.4 & 1582.7 & 1583.9 & 1713.9 & \textbf{1683.1} & 1697.6 & 1645.1 & 72.0\% \\
3 & claude-opus-4.6 & Baseline & 1615.5 & 1617.3 & \textbf{1662.1} & 1646.4 & 1664.7 & 1654.4 & 1643.4 & 71.8\% \\
4 & gpt-5.4 & Agent & 1607.0 & 1628.6 & 1603.1 & 1645.1 & 1651.8 & 1638.7 & 1629.0 & 73.1\% \\
5 & gpt-5.4 & Baseline & \textbf{1627.3} & \textbf{1638.2} & 1604.8 & 1623.0 & 1630.8 & 1611.3 & 1622.6 & 71.0\% \\
6 & glm-5.1 & Agent & 1542.1 & 1562.4 & 1580.0 & 1649.5 & 1601.3 & 1649.5 & 1597.5 & 60.2\% \\
7 & kimi-k2.6 & Agent & 1553.3 & 1570.7 & 1552.6 & 1601.8 & 1611.4 & 1629.9 & 1586.6 & 59.7\% \\
\rowcolor{colorGT} 8 & reference &  & 1578.9 & 1565.6 & 1516.0 & 1615.3 & 1598.1 & 1645.6 & 1586.6 & 59.9\% \\
9 & claude-opus-4.6 & Agent & 1540.7 & 1518.6 & 1533.2 & 1628.4 & 1647.7 & 1645.4 & 1585.7 & 57.1\% \\
10 & glm-5.1 & Baseline & 1597.6 & 1595.0 & 1594.5 & 1554.1 & 1582.0 & 1547.0 & 1578.4 & 60.0\% \\
11 & kimi-k2.6 & Baseline & 1580.0 & 1566.4 & 1565.8 & 1585.7 & 1560.3 & 1609.4 & 1577.9 & 59.2\% \\
12 & deepseek-v4-pro & Baseline & 1543.7 & 1528.6 & 1573.3 & 1525.7 & 1510.5 & 1538.2 & 1536.7 & 48.0\% \\
13 & qwen-3.6-max & Baseline & 1519.0 & 1525.3 & 1516.1 & 1508.2 & 1493.9 & 1503.1 & 1510.9 & 44.3\% \\
14 & qwen-3.6-max & Agent & 1502.9 & 1506.5 & 1495.6 & 1527.3 & 1518.4 & 1512.3 & 1510.5 & 42.2\% \\
15 & minimax-m2.7 & Agent & 1492.2 & 1485.6 & 1481.0 & 1534.0 & 1497.6 & 1535.1 & 1504.3 & 39.0\% \\
16 & minimax-m2.7 & Baseline & 1526.5 & 1515.6 & 1509.2 & 1515.4 & 1509.7 & 1432.5 & 1501.5 & 39.7\% \\
17 & deepseek-v4-pro & Agent & 1480.5 & 1504.7 & 1499.3 & 1475.0 & 1483.1 & 1514.6 & 1492.9 & 38.3\% \\
18 & deepseek-v4-flash & Baseline & 1482.8 & 1529.6 & 1525.5 & 1488.6 & 1427.6 & 1475.9 & 1488.3 & 38.2\% \\
19 & glm-5 & Agent & 1462.3 & 1479.0 & 1476.9 & 1485.2 & 1501.7 & 1484.5 & 1481.6 & 36.0\% \\
20 & glm-5 & Baseline & 1496.2 & 1469.3 & 1474.7 & 1480.7 & 1481.8 & 1443.9 & 1474.4 & 36.1\% \\
21 & gpt-5.4-mini & Agent & 1455.8 & 1463.4 & 1436.0 & 1493.0 & 1452.0 & 1466.1 & 1461.1 & 32.2\% \\
22 & deepseek-v4-flash & Agent & 1417.1 & 1451.1 & 1463.9 & 1446.6 & 1427.1 & 1488.0 & 1449.0 & 29.8\% \\
23 & gpt-5.4-mini & Baseline & 1448.4 & 1458.8 & 1416.8 & 1446.5 & 1409.1 & 1428.1 & 1434.6 & 27.3\% \\
24 & minimax-m2.5 & Baseline & 1481.3 & 1444.3 & 1453.5 & 1422.6 & 1435.3 & 1360.5 & 1432.9 & 26.4\% \\
25 & minimax-m2.5 & Agent & 1446.1 & 1415.5 & 1397.7 & 1445.3 & 1458.2 & 1401.8 & 1427.5 & 23.9\% \\
26 & gemini-3.1-pro & Agent & 1443.6 & 1427.1 & 1418.2 & 1354.7 & 1413.6 & 1409.6 & 1411.1 & 23.6\% \\
27 & gemini-3.1-pro & Baseline & 1457.3 & 1428.3 & 1440.5 & 1338.9 & 1389.0 & 1377.9 & 1405.3 & 24.2\% \\
28 & kimi-k2.5 & Agent & 1333.8 & 1351.4 & 1380.8 & 1465.3 & 1369.8 & 1475.8 & 1396.1 & 21.1\% \\
29 & gemini-3-flash & Baseline & 1445.3 & 1441.3 & 1439.3 & 1257.9 & 1301.4 & 1262.5 & 1357.9 & 18.0\% \\
30 & gemini-3-flash & Agent & 1397.3 & 1408.0 & 1415.4 & 1215.4 & 1305.2 & 1238.2 & 1329.9 & 13.2\% \\
31 & kimi-k2.5 & Baseline & 1328.1 & 1285.1 & 1355.2 & 1258.6 & 1337.8 & 1282.5 & 1307.9 & 11.9\% \\
\bottomrule
\end{tabular}
\end{table}

\begin{table}[t]
\centering
\caption{Rubric scoring diagnostics under \texttt{seed-2.0-pro}: per-(model, mode) mean rubric $S$ on the 1--5 scale, complementing Figure~\ref{fig:arena-vs-score} of Section~\ref{subsec:arena-vs-score} and Table~\ref{tab:arena-all-seed}. Rows are sorted by arena rank (Arena \#) on the same full-pool scale as that table; \textit{Rubric Rank} is the position when the same cells are re-ranked by Avg $S$ (descending), with $\Delta = $ Rubric Rank $-$ Arena Rank. The narrow Avg $S$ spread and frequent non-zero $\Delta$ instantiate the compression and protocol-dependent reordering analyzed in Section~\ref{subsec:arena-vs-score}.}
\label{tab:score-seed}
\small
\setlength{\tabcolsep}{2.5pt}
\begin{tabular}{cll rrr rrr rc}
\toprule
 & & & \multicolumn{3}{c}{\textbf{Scientific Domains}} & \multicolumn{3}{c}{\textbf{Analytical Domains}} & & \\
\cmidrule(lr){4-6} \cmidrule(lr){7-9}
\textbf{Arena \#} & \textbf{Model} & \textbf{Mode} & \textbf{Bio} & \textbf{ML} & \textbf{Social} & \textbf{Fin} & \textbf{IT} & \textbf{Safety} & \textbf{Avg $S$} & \textbf{Rubric Rank ($\Delta$)} \\
\midrule
1 & claude-sonnet-4.6 & Agent & 4.70 & 4.92 & 4.89 & 4.77 & 4.37 & 4.71 & 4.73 & 4~{\scriptsize ($\Delta$+3)} \\
2 & claude-sonnet-4.6 & Baseline & 4.79 & 4.92 & 4.89 & 4.54 & 4.40 & 4.62 & 4.69 & 5~{\scriptsize ($\Delta$+3)} \\
3 & claude-opus-4.6 & Baseline & 4.76 & 4.89 & 4.90 & 4.25 & 4.29 & 4.57 & 4.61 & 6~{\scriptsize ($\Delta$+3)} \\
4 & gpt-5.4 & Agent & 4.75 & 4.95 & 4.91 & 4.63 & 4.46 & 4.69 & 4.73 & 3~{\scriptsize ($\Delta$-1)} \\
5 & gpt-5.4 & Baseline & 4.76 & 4.96 & 4.91 & 4.57 & 4.53 & 4.66 & 4.73 & 2~{\scriptsize ($\Delta$-3)} \\
6 & glm-5.1 & Agent & 4.57 & 4.76 & 4.77 & 4.29 & 4.24 & 4.52 & 4.53 & 10~{\scriptsize ($\Delta$+4)} \\
7 & kimi-k2.6 & Agent & 4.59 & 4.79 & 4.71 & 4.23 & 4.25 & 4.45 & 4.50 & 12~{\scriptsize ($\Delta$+5)} \\
\rowcolor{colorGT} 8 & reference &  & 4.76 & 4.88 & 4.75 & 4.95 & 4.86 & 4.93 & 4.85 & 1~{\scriptsize ($\Delta$-7)} \\
9 & claude-opus-4.6 & Agent & 4.62 & 4.81 & 4.78 & 4.20 & 4.21 & 4.58 & 4.53 & 9~{\scriptsize ($\Delta$0)} \\
10 & glm-5.1 & Baseline & 4.75 & 4.86 & 4.82 & 4.25 & 4.27 & 4.41 & 4.56 & 8~{\scriptsize ($\Delta$-2)} \\
11 & kimi-k2.6 & Baseline & 4.67 & 4.85 & 4.83 & 4.37 & 4.25 & 4.56 & 4.59 & 7~{\scriptsize ($\Delta$-4)} \\
12 & deepseek-v4-pro & Baseline & 4.45 & 4.67 & 4.62 & 4.04 & 4.00 & 4.36 & 4.36 & 18~{\scriptsize ($\Delta$+6)} \\
13 & qwen-3.6-max & Baseline & 4.55 & 4.81 & 4.74 & 4.23 & 4.25 & 4.49 & 4.51 & 11~{\scriptsize ($\Delta$-2)} \\
14 & qwen-3.6-max & Agent & 4.51 & 4.75 & 4.70 & 4.17 & 4.16 & 4.40 & 4.45 & 15~{\scriptsize ($\Delta$+1)} \\
15 & minimax-m2.7 & Agent & 4.44 & 4.62 & 4.59 & 3.93 & 3.92 & 4.20 & 4.28 & 21~{\scriptsize ($\Delta$+6)} \\
16 & minimax-m2.7 & Baseline & 4.60 & 4.75 & 4.72 & 4.09 & 3.96 & 4.16 & 4.38 & 17~{\scriptsize ($\Delta$+1)} \\
17 & deepseek-v4-pro & Agent & 4.28 & 4.56 & 4.48 & 3.87 & 3.94 & 4.22 & 4.23 & 24~{\scriptsize ($\Delta$+7)} \\
18 & deepseek-v4-flash & Baseline & 4.26 & 4.57 & 4.54 & 4.02 & 3.80 & 4.28 & 4.24 & 22~{\scriptsize ($\Delta$+4)} \\
19 & glm-5 & Agent & 4.37 & 4.57 & 4.61 & 4.07 & 4.08 & 4.28 & 4.33 & 19~{\scriptsize ($\Delta$0)} \\
20 & glm-5 & Baseline & 4.52 & 4.72 & 4.68 & 4.16 & 4.20 & 4.38 & 4.44 & 16~{\scriptsize ($\Delta$-4)} \\
21 & gpt-5.4-mini & Agent & 4.46 & 4.76 & 4.63 & 4.35 & 4.17 & 4.41 & 4.46 & 14~{\scriptsize ($\Delta$-7)} \\
22 & deepseek-v4-flash & Agent & 4.10 & 4.44 & 4.45 & 3.92 & 3.83 & 4.29 & 4.17 & 25~{\scriptsize ($\Delta$+3)} \\
23 & gpt-5.4-mini & Baseline & 4.48 & 4.76 & 4.62 & 4.34 & 4.14 & 4.44 & 4.46 & 13~{\scriptsize ($\Delta$-10)} \\
24 & minimax-m2.5 & Baseline & 4.44 & 4.63 & 4.59 & 3.85 & 3.95 & 4.01 & 4.24 & 23~{\scriptsize ($\Delta$-1)} \\
25 & minimax-m2.5 & Agent & 4.21 & 4.43 & 4.32 & 3.88 & 3.85 & 4.07 & 4.13 & 28~{\scriptsize ($\Delta$+3)} \\
26 & gemini-3.1-pro & Agent & 4.27 & 4.47 & 4.45 & 3.68 & 3.93 & 4.20 & 4.17 & 26~{\scriptsize ($\Delta$0)} \\
27 & gemini-3.1-pro & Baseline & 4.33 & 4.48 & 4.46 & 3.96 & 4.13 & 4.36 & 4.29 & 20~{\scriptsize ($\Delta$-7)} \\
28 & kimi-k2.5 & Agent & 4.01 & 4.31 & 4.28 & 3.73 & 3.79 & 4.02 & 4.02 & 30~{\scriptsize ($\Delta$+2)} \\
29 & gemini-3-flash & Baseline & 4.17 & 4.42 & 4.35 & 3.70 & 3.99 & 4.16 & 4.13 & 27~{\scriptsize ($\Delta$-2)} \\
30 & gemini-3-flash & Agent & 4.00 & 4.33 & 4.30 & 3.51 & 3.80 & 4.03 & 4.00 & 31~{\scriptsize ($\Delta$+1)} \\
31 & kimi-k2.5 & Baseline & 3.99 & 4.31 & 4.34 & 4.01 & 3.84 & 4.01 & 4.08 & 29~{\scriptsize ($\Delta$-2)} \\
\bottomrule
\end{tabular}
\end{table}

\begin{table}[t]
\centering
\caption{Per-rubric-dimension breakdown for \textbf{Financial Analysis} under \texttt{seed-2.0-pro} (full pool). Rows sorted by Avg $S$ descending; all dimensions on the 1--5 scale.}
\label{tab:score-details-fin-seed}
\small
\setlength{\tabcolsep}{4pt}
\begin{tabular}{cll rrrrrrr}
\toprule
\textbf{\#} & \textbf{Model} & \textbf{Mode} & \textbf{Ground.} & \textbf{Insight} & \textbf{Justif.} & \textbf{Breadth} & \textbf{Distinct.} & \textbf{Utility} & \textbf{Avg $S$} \\
\midrule
\rowcolor{colorGT} 1 & reference &  & 4.99 & 4.71 & 4.98 & 5.00 & 5.00 & 5.00 & 4.95 \\
2 & claude-sonnet-4.6 & Agent & 5.00 & 4.25 & 4.88 & 4.50 & 5.00 & 5.00 & 4.77 \\
3 & gpt-5.4 & Agent & 4.77 & 4.13 & 4.66 & 4.46 & 4.92 & 4.88 & 4.63 \\
4 & gpt-5.4 & Baseline & 4.75 & 4.06 & 4.67 & 4.39 & 4.77 & 4.79 & 4.57 \\
5 & claude-sonnet-4.6 & Baseline & 4.61 & 4.09 & 4.51 & 4.55 & 4.82 & 4.68 & 4.54 \\
6 & kimi-k2.6 & Baseline & 4.55 & 3.90 & 4.41 & 4.22 & 4.73 & 4.38 & 4.37 \\
7 & gpt-5.4-mini & Agent & 4.61 & 3.81 & 4.49 & 4.11 & 4.68 & 4.39 & 4.35 \\
8 & gpt-5.4-mini & Baseline & 4.62 & 3.81 & 4.51 & 4.11 & 4.68 & 4.34 & 4.34 \\
9 & glm-5.1 & Agent & 4.37 & 3.94 & 4.29 & 4.16 & 4.54 & 4.41 & 4.29 \\
10 & claude-opus-4.6 & Baseline & 4.28 & 3.89 & 4.15 & 4.23 & 4.61 & 4.37 & 4.25 \\
11 & glm-5.1 & Baseline & 4.41 & 3.84 & 4.30 & 4.15 & 4.64 & 4.27 & 4.25 \\
12 & qwen-3.6-max & Baseline & 4.31 & 3.85 & 4.27 & 4.04 & 4.65 & 4.25 & 4.23 \\
13 & kimi-k2.6 & Agent & 4.23 & 3.88 & 4.13 & 4.20 & 4.61 & 4.32 & 4.23 \\
14 & claude-opus-4.6 & Agent & 4.17 & 3.82 & 4.08 & 4.24 & 4.56 & 4.32 & 4.20 \\
15 & qwen-3.6-max & Agent & 4.17 & 3.89 & 4.09 & 3.99 & 4.61 & 4.26 & 4.17 \\
16 & glm-5 & Baseline & 4.39 & 3.77 & 4.29 & 3.91 & 4.50 & 4.13 & 4.16 \\
17 & minimax-m2.7 & Baseline & 4.21 & 3.66 & 4.04 & 4.05 & 4.52 & 4.06 & 4.09 \\
18 & glm-5 & Agent & 4.17 & 3.74 & 4.06 & 3.96 & 4.46 & 4.04 & 4.07 \\
19 & deepseek-v4-pro & Baseline & 4.11 & 3.73 & 3.94 & 4.07 & 4.44 & 4.04 & 4.04 \\
20 & deepseek-v4-flash & Baseline & 4.06 & 3.74 & 3.88 & 4.02 & 4.38 & 4.08 & 4.02 \\
21 & kimi-k2.5 & Baseline & 4.52 & 3.09 & 4.37 & 4.19 & 4.21 & 3.61 & 4.01 \\
22 & gemini-3.1-pro & Baseline & 4.03 & 3.64 & 3.90 & 3.82 & 4.45 & 3.90 & 3.96 \\
23 & minimax-m2.7 & Agent & 3.95 & 3.55 & 3.89 & 4.00 & 4.27 & 3.91 & 3.93 \\
24 & deepseek-v4-flash & Agent & 3.91 & 3.71 & 3.74 & 3.93 & 4.33 & 3.92 & 3.92 \\
25 & minimax-m2.5 & Agent & 3.88 & 3.49 & 3.74 & 3.95 & 4.25 & 3.95 & 3.88 \\
26 & deepseek-v4-pro & Agent & 3.80 & 3.63 & 3.70 & 3.89 & 4.34 & 3.86 & 3.87 \\
27 & minimax-m2.5 & Baseline & 3.92 & 3.42 & 3.74 & 3.98 & 4.15 & 3.88 & 3.85 \\
28 & kimi-k2.5 & Agent & 3.66 & 3.40 & 3.51 & 4.08 & 3.98 & 3.79 & 3.73 \\
29 & gemini-3-flash & Baseline & 3.70 & 3.45 & 3.48 & 3.78 & 4.22 & 3.61 & 3.70 \\
30 & gemini-3.1-pro & Agent & 3.43 & 3.53 & 3.31 & 3.82 & 4.32 & 3.68 & 3.68 \\
31 & gemini-3-flash & Agent & 3.33 & 3.33 & 3.14 & 3.68 & 4.24 & 3.37 & 3.51 \\
\bottomrule
\end{tabular}
\end{table}

Three regularities emerge directly from the comparison.

\textit{Compression.} Every Avg $S$ falls within a single rubric point; the strongest and weakest cells in the paper's evaluation set are separated by less than the distance between $4.0$ and $5.0$, while the same cells span several hundred BTD points under arena. The compression is visible in every domain column.

\textit{Ceiling effect.} Almost every row sits above $4.0$, with the majority above $4.3$. Rubric judges default to near-top scores for anything not manifestly deficient, and this behaviour concentrates at the top of the scale rather than the middle. The concentration is structural: absent a single canonical answer, committing to a low absolute score is harder than declaring one of two submissions better in head-to-head comparison.

\textit{Protocol-dependent reordering.} Relative to the arena ordering, 28 of 31 full-pool entries receive a different rubric rank. The top three arena entries each rank three positions lower under rubric scoring, while the Reference moves from rank 8 under arena to rank 1 under rubric. These shifts demonstrate substantial protocol sensitivity in fine-grained ordering, but they do not identify the underlying mechanism.

Together, score compression, ceiling effects, and widespread reordering support using arena BTD for primary comparative ranking and rubric scores for diagnostic analysis.

\section{Extended Analysis of Evaluator Alignment}
\label{app:alignment}

We examine two validation signals that complement the cross-judge analysis in Appendix~\ref{app:judge-triangulation}. Human preference judgments assess alignment between automated rankings and expert preferences, while ICLR acceptance outcomes provide a source-side signal not used by the arena judge. Because these analyses operate at different granularities, we interpret each within the scope of the evidence it provides.

\subsection{Detailed Analysis of Human-Judge Alignment}
\label{app:arena-judge-validation}

To assess alignment between the primary arena judge and expert preferences, we compare rankings induced by \texttt{seed-2.0-pro} with rankings aggregated from 1,500 expert pairwise judgments across Financial Analysis, Social Science, and Biomedical Science. Detailed annotation procedures are provided in Appendix~\ref{app:preference-evaluation}.

\begin{table}[t]
\centering
\caption{Per-domain leaderboard agreement between human experts and the \texttt{seed-2.0-pro} judge across 16 models. $\Delta$ is the rank shift from human to judge; per-domain Kendall's $\tau$ and Spearman's $\rho$ are reported at the bottom. Most models show small displacements, with the largest spread concentrated in Biomedical Science.}
\label{tab:arena-per-domain}
\small
\setlength{\tabcolsep}{5pt} 
\begin{tabular}{@{} l ccc ccc ccc @{}}
\toprule
\multirow{2.5}{*}{\textbf{Model}} & \multicolumn{3}{c}{\textbf{Biomedical}} & \multicolumn{3}{c}{\textbf{Financial Analysis}} & \multicolumn{3}{c}{\textbf{Social Science}} \\
\cmidrule(lr){2-4} \cmidrule(lr){5-7} \cmidrule(lr){8-10}
 & Rank & Win\% & $\Delta$ & Rank & Win\% & $\Delta$ & Rank & Win\% & $\Delta$ \\
\midrule
claude-sonnet-4.6-high & \underline{2} & 90.9 & 0  & \textbf{1} & 97.5 & 0  & \textbf{1} & 93.8 & 0  \\
gpt-5.4-high           & \textbf{1} & 81.8 & 0  & 4 & 88.5 & 0  & \underline{2} & 71.4 & 0  \\
claude-opus-4.6-high   & \underline{3} & 86.7 & 0  & 5 & 73.5 & 0  & 4 & 75.0 & -1 \\
qwen-3.6-max-thinking  & 8 & 44.4 & +4 & \underline{2} & 75.8 & 0  & 5 & 71.7 & 0  \\
kimi-k2.6-thinking     & 5 & 66.7 & +1 & \underline{3} & 81.2 & 0  & 7 & 58.3 & -1 \\
glm-5.1-thinking       & 4 & 70.6 & 0  & 6 & 63.4 & 0  & \underline{3} & 73.9 & +1 \\
qwen-3.6-plus-thinking & 6 & 57.1 & +9 & 8 & 54.8 & 0  & 6 & 64.3 & +2 \\
glm-5-thinking         & 15& 18.2 & -6 & 7 & 63.9 & 0  & 14& 23.1 & -3 \\
gemini-3.1-pro-high    & 7 & 54.5 & 0  & 12& 23.5 & +2 & 9 & 51.9 & -2 \\
kimi-k2.5-thinking     & 14& 28.6 & -1 & 9 & 44.3 & 0  & 11& 43.2 & +1 \\
gpt-5.4-mini-high      & 12& 37.5 & -7 & 10& 40.0 & 0  & 13& 26.0 & 0  \\
minimax-m2.7-thinking  & 11& 38.5 & -3 & 11& 29.0 & 0  & 10& 51.9 & 0  \\
deepseek-v4-flash-high & 10& 40.0 & +4 & 13& 20.0 & -1 & 16& 11.3 & -1 \\
minimax-m2.5-thinking  & 9 & 44.4 & +2 & 14& 18.8 & -1 & 8 & 50.0 & +1 \\
gemini-3-flash-high    & 13& 30.0 & -3 & 15& 2.9  & 0  & 15& 17.3 & +1 \\
deepseek-v4-pro-high   & 16& 5.9  & 0  & 16& 0.0  & 0  & 12& 42.0 & +2 \\
\midrule
\multicolumn{10}{@{}l}{\textbf{Correlations} (Kendall's $\tau$ / Spearman's $\rho$)} \\
\multicolumn{10}{@{}l}{Biomedical: 0.53 / 0.67 \quad Financial: 0.97 / 0.99 \quad Social Science: 0.85 / 0.96} \\
\bottomrule
\end{tabular}
\end{table}

As shown in Table~\ref{tab:arena-per-domain}, agreement is strong but varies by domain. The top three positions match exactly in Biomedical Science and Financial Analysis, whereas Social Science differs by a swap between ranks 3 and 4. Most rank shifts are small, although Biomedical Science contains larger displacements among lower-ranked models, including $\Delta=+9$ for \texttt{qwen-3.6-plus} and $\Delta=-7$ for \texttt{gpt-5.4-mini}.

\paragraph{Lower alignment in Biomedical Science.}
The lower correlation in Biomedical Science ($\tau=0.53$) indicates greater evaluator disagreement than in Financial Analysis ($\tau=0.97$). The current analysis does not isolate the source of this difference, so we do not attribute it to a particular property of the cases or annotators. Nevertheless, the matching top-three positions in Biomedical Science and the concentration of the largest shifts among lower-ranked models show that disagreement does not extend uniformly across the leaderboard. The stronger correlations in Financial Analysis and Social Science further indicate that alignment varies by domain.

\subsection{Association with Peer-Review Outcomes}
\label{subapp:alignment-ml}

The Machine Learning source papers were sampled from ICLR 2026 and stratified by acceptance category, providing an external outcome not used by the arena judge. If arena evaluation is sensitive to source-side differences associated with peer-review outcomes, the source-derived Reference constructed by Forge from each source paper should be more competitive on accepted cases than on rejected ones.

\begin{table}[t]
\centering
\caption{Reference arena strength under \texttt{seed-2.0-pro} (baseline pool) on the \texttt{machine\_learning} subset, split by ICLR 2026 acceptance. \textit{Debiased}: per-case median of Reference's $[0,1]$ debiased win share. \textit{Win Rate}: per-case median of the AND-of-both win indicator, where Reference wins both forward and reverse, and ties score $0.5$. \textit{BTD}: Reference's BTD rating fit on the bucket's matches alone.}
\label{tab:alignment-ml}
\small
\setlength{\tabcolsep}{8pt}
\begin{tabular}{lccc}
\toprule
\textbf{Group} & \textbf{Debiased ($\uparrow$)} & \textbf{Win Rate ($\uparrow$)} & \textbf{BTD ($\uparrow$)} \\
\midrule
Accepted & 0.602 & 0.654 & 1577.6 \\
Rejected & 0.538 & 0.615 & 1530.4 \\
\bottomrule
\end{tabular}
\end{table}

Table~\ref{tab:alignment-ml} shows a consistent directional pattern across all three measurements. Accepted cases exceed rejected cases by $0.06$ in median per-case debiased score and $0.04$ in median per-case win rate, while the bucket-specific BTD rating is 47 points higher. This agreement provides a complementary external check using information unavailable to the arena judge. Because acceptance reflects factors beyond hypothesis quality, we do not treat it as case-level ground truth.

\section{Inter-Judge Agreement Analysis}
\label{app:judge-triangulation}

The headline analyses in Section~\ref{sec:experiments} use \texttt{seed-2.0-pro} as the sole arena judge. To test whether the resulting rankings are an artifact of that single model's idiosyncrasies, we re-run the entire arena pipeline with an independent second judge, \texttt{mimo-v2-pro}, on the same submissions and the same case set. This appendix reports five complementary agreement signals between the two judges, spanning three granularities.

\paragraph{Three levels of agreement.}
We measure cross-judge agreement at three granularities, all restricted to the baseline-pool model set used for the main leaderboard:
\begin{itemize}\itemsep=2pt
    \item \textit{Full-ranking agreement}: per-domain Spearman $\rho$ and Kendall $\tau$ between the two judges' BTD ratings (one value per (model, domain) cell), and \textit{Mean Rank Shift}, the average $|\Delta\text{rank}|$ per model when switching judges.
    \item \textit{Top-tier identity}: per-domain \textit{Top-5 Overlap}, i.e.\ the size of the intersection of the two judges' top-5 ranked models.
    \item \textit{Individual verdicts}: per-domain \textit{Pairwise Agreement}, i.e.\ the fraction of shared (case, model$_a$, model$_b$) matches where both judges' forward verdicts select the same outcome among A wins, B wins, and tie.
\end{itemize}
Table~\ref{tab:arena-arena} reports all five statistics per domain plus a pooled row across all (model $\times$ domain) cells.

\begin{table}[t]
\centering
\caption{Inter-judge agreement between \texttt{seed-2.0-pro} and \texttt{mimo-v2-pro} on the 16 baseline-mode submissions evaluated by both judges in each domain. Spearman $\rho$ and Kendall $\tau$ are rank correlations between the two judges' BTD ratings; Rank Shift is the mean $|\Delta\text{rank}|$ per model; Top-5 is the intersection size of each judge's top-5 models; Pairwise is the fraction of shared (case, model$_a$, model$_b$) matches where both judges' forward verdicts agree.}
\label{tab:arena-arena}
\small
\setlength{\tabcolsep}{6pt}
\begin{tabular}{lccccc}
\toprule
\textbf{Domain} & \textbf{Spearman $\rho$ ($\uparrow$)} & \textbf{Kendall $\tau$ ($\uparrow$)} & \textbf{Rank Shift ($\downarrow$)} & \textbf{Top-5 ($\uparrow$)} & \textbf{Pairwise ($\uparrow$)} \\
\midrule
Biomedical Science & 0.915 & 0.750 & 1.50 & 5/5 & 63.3\% \\
Machine Learning & 0.950 & 0.833 & 1.12 & 5/5 & 67.0\% \\
Social Science & 0.974 & 0.900 & 0.75 & 5/5 & 66.8\% \\
Financial Analysis & 0.909 & 0.833 & 1.00 & 4/5 & 68.1\% \\
IT Operations & 0.900 & 0.783 & 1.50 & 4/5 & 67.5\% \\
Safety Investigation & 0.944 & 0.817 & 1.12 & 4/5 & 65.6\% \\
\midrule
\textbf{Overall} & \textbf{0.920} & \textbf{0.764} & \textbf{1.17} & \textbf{4.50/5} & \textbf{66.3\%} \\
\bottomrule
\end{tabular}
\end{table}

\paragraph{Rankings replicate; single-match verdicts are noisier.}
The two judges produce essentially the same leaderboard. Per-domain rank correlations are uniformly strong (Spearman $\rho \in [0.90, 0.97]$; Kendall $\tau \in [0.75, 0.90]$), the pooled $\rho = 0.920$ has a tight bootstrap 95\% CI of $[0.871, 0.949]$, and the \textit{Mean Rank Shift} of $1.17$ on a 16-model scale translates this into concrete terms: switching judges moves a model's leaderboard position by only about one place on average. \textit{Top-5 Overlap} averages $4.5/5$ across domains, with three domains in perfect agreement and the remaining three differing by a single model. The top tier's identity is robust to judge choice. Pairwise verdict agreement ranges from $63\%$ to $68\%$ across domains, indicating substantial but incomplete agreement at the individual-match level. Despite disagreement on roughly one third of individual matches, the BTD-aggregated rankings remain closely aligned across judges. This rank-level stability supports the use of arena aggregation for the primary leaderboard.

\begin{table}[t]
\centering
\caption{Main leaderboard under reference-independent pairwise judging by \texttt{mimo-v2-pro} as the independent second judge (companion to Table~\ref{tab:arena-seed}). All BTD and WR conventions match Table~\ref{tab:arena-seed}; Reference participates as an anonymous competitor for calibration.}
\label{tab:arena-mimo}
\small
\setlength{\tabcolsep}{3.5pt}
\begin{tabular}{cll rrr rrr rr}
\toprule
 & & & \multicolumn{3}{c}{\textbf{Scientific Domains}} & \multicolumn{3}{c}{\textbf{Analytical Domains}} & & \\
\cmidrule(lr){4-6} \cmidrule(lr){7-9}
\textbf{\#} & \textbf{Model} & \textbf{Effort} & \textbf{Bio} & \textbf{ML} & \textbf{Social} & \textbf{Fin} & \textbf{IT} & \textbf{Safety} & \textbf{Avg} & \textbf{WR} \\
\midrule
1 & claude-sonnet-4.6 & high & \textbf{1642.3} & 1618.9 & 1589.3 & \textbf{1640.2} & \textbf{1632.3} & \textbf{1609.9} & \textbf{1622.2} & 67.7\% \\
2 & claude-opus-4.6 & high & 1632.2 & \textbf{1634.7} & \textbf{1620.2} & 1610.7 & 1589.4 & 1574.3 & 1610.3 & 62.6\% \\
3 & gpt-5.4 & high & 1578.2 & 1602.9 & 1570.0 & 1556.0 & 1548.3 & 1542.1 & 1566.3 & 49.1\% \\
4 & glm-5.1 & \checkmark & 1583.9 & 1593.2 & 1562.7 & 1549.4 & 1543.2 & 1543.8 & 1562.7 & 49.0\% \\
5 & kimi-k2.6 & \checkmark & 1530.3 & 1547.1 & 1542.8 & 1555.7 & 1549.7 & 1569.3 & 1549.1 & 44.3\% \\
6 & deepseek-v4-pro & high & 1514.0 & 1519.3 & 1556.1 & 1535.6 & 1534.0 & 1547.7 & 1534.5 & 41.5\% \\
\rowcolor{colorGT} 7 & reference &  & 1494.7 & 1538.0 & 1485.8 & 1558.5 & 1532.1 & 1567.2 & 1529.4 & 39.3\% \\
8 & deepseek-v4-flash & high & 1489.0 & 1505.8 & 1513.4 & 1564.3 & 1533.5 & 1548.7 & 1525.8 & 40.0\% \\
9 & qwen-3.6-max & \checkmark & 1519.8 & 1520.4 & 1517.6 & 1523.3 & 1499.7 & 1509.7 & 1515.1 & 35.1\% \\
10 & minimax-m2.7 & \checkmark & 1487.9 & 1484.0 & 1461.8 & 1509.9 & 1514.4 & 1470.8 & 1488.1 & 26.9\% \\
11 & glm-5 & \checkmark & 1467.6 & 1456.6 & 1452.6 & 1496.2 & 1482.0 & 1488.2 & 1473.9 & 25.2\% \\
12 & gpt-5.4-mini & high & 1475.8 & 1422.8 & 1429.3 & 1461.1 & 1496.3 & 1458.4 & 1457.3 & 20.2\% \\
13 & gemini-3.1-pro & high & 1438.6 & 1426.3 & 1448.0 & 1442.7 & 1426.2 & 1457.5 & 1439.9 & 19.7\% \\
14 & minimax-m2.5 & \checkmark & 1423.3 & 1412.6 & 1434.4 & 1451.8 & 1436.7 & 1435.2 & 1432.3 & 15.1\% \\
15 & gemini-3-flash & high & 1432.7 & 1436.4 & 1455.2 & 1373.9 & 1386.8 & 1385.8 & 1411.8 & 16.0\% \\
16 & kimi-k2.5 & \checkmark & 1310.3 & 1300.0 & 1383.4 & 1193.8 & 1324.8 & 1320.6 & 1305.5 & 7.2\% \\
\bottomrule
\end{tabular}
\end{table}

Table~\ref{tab:arena-mimo} reports the baseline leaderboard under \texttt{mimo-v2-pro} as a companion to Table~\ref{tab:arena-seed}. Both judges rank \texttt{claude-sonnet-4.6} first, \texttt{claude-opus-4.6} second, and \texttt{gpt-5.4} third, with \texttt{kimi-k2.5} last. They also yield the same within-family ordering for the DeepSeek, GLM, and Kimi model pairs.

Fine-grained disagreement is concentrated in the middle of the leaderboard. The Reference moves from rank 4 under \texttt{seed-2.0-pro} to rank 7 under \texttt{mimo-v2-pro}, the largest displacement, while no other system moves by more than two positions. Because this analysis does not isolate the source of judge disagreement, we interpret the Reference shift as judge sensitivity in its relative placement rather than evidence of a specific stylistic mechanism. The mean rank shift of $1.17$ and pooled Spearman $\rho=0.920$ with 95\% CI $[0.871, 0.949]$ show that the overall ranking structure remains closely aligned.

\paragraph{Complementary validation signals.}
We consider cross-judge agreement alongside two additional checks. The first measures alignment with human expert preferences in three domains (\S\ref{subsec:alignment}, Kendall $\tau=0.90$), and the second examines the directional association between the competitiveness of the source-derived Reference and ICLR acceptance outcome (Appendix~\ref{subapp:alignment-ml}, with a 47-point higher bucket-specific BTD rating on accepted cases). These analyses operate at different granularities and do not constitute three replications of the same leaderboard. Instead, they probe sensitivity to judge choice, expert alignment, and association with a peer-review outcome not used by the arena judge. Together, they broaden the evidence base for \textsc{HypoEval} without treating any single signal as definitive.

\section{Prompt Templates}
\label{app:prompts}

\newcommand{\phead}[1]{{\fontfamily{lmtt}\bfseries\selectfont #1}}

This appendix provides the complete prompt templates used in the \textsc{HypoArena} pipeline for two representative domains: Social Science (scientific research, single-hypothesis) and Safety Investigation (analytical investigation, multi-hypothesis). The full set for all six domains is released with our code.

\subsection{Stage 1: Context Construction}

\subsubsection{Context Forge}

\begin{tcolorbox}[colback=gray!5, colframe=gray!40, coltitle=black, fonttitle=\small\bfseries, fontupper=\scriptsize\ttfamily, title={Context Forge --- Social Science}, breakable]
You are ForgeAgent. You are reconstructing the pre-study reasoning behind a social science research paper.\\[4pt]
This is an iterative process: AuditAgent will review your output and may send feedback asking you to revise. When that happens, use the full conversation history to understand what you previously produced and why, then make targeted revisions rather than starting from scratch.\\[4pt]
Think like a benchmark constructor, not like an author writing a paper introduction. A strong Context feels like the background and unresolved-tension section of a serious survey --- a reader comes away thinking ``there is enough background and tension here to support a serious hypothesis, but the answer has not been given away.''\\[4pt]
You have access to the original source document and web search.\\[4pt]
\phead{Domain Guidance}\\
- Cover the relevant theories, constructs, operationalizations, competing mechanisms, and empirical traditions. End at the unresolved choice among mechanisms or trade-offs, not at the preferred mediator.\\
- Do not name the paper's preferred mediator, moderator, or framing variable as the explanation.\\
- Actively use search to enrich the context beyond the paper's local introduction. Prefer reliable and substantive sources.\\
- Write a technical survey-style context that integrates paper background with external primary-source background.\\[4pt]
\phead{How to Approach This Task}\\
- Read the source document to identify the problem setting, factual background, prior approaches, existing evidence, and the tensions it surfaces.\\
- Search is a thinking tool for widening the frame. A good self-check: could this draft have been written from the paper alone? If yes, it probably needs search-informed widening.\\
- Go deep on the tensions that matter rather than cataloguing everything in shallow form.\\
- End at unresolved tension, disagreement, vulnerability, or missing evidence, not at a solution-shaped landing.\\
- Keep the final text clean of citations, URLs, bibliography, or any retrieval residue.\\[4pt]
\phead{Self-Check}: (1) Did you use search to widen? (2) Field-level background, not rewritten Introduction? (3) Could a reader infer the answer? (4) Ends at tension? (5) Sufficient density?\\[4pt]
\phead{Output}: \{``context'': ``<The benchmark context>''\}
\end{tcolorbox}

\begin{tcolorbox}[colback=gray!5, colframe=gray!40, coltitle=black, fonttitle=\small\bfseries, fontupper=\scriptsize\ttfamily, title={Context Forge --- Safety Investigation}, breakable]
You are ForgeAgent. Your job is to construct model-visible benchmark Context from an official accident or incident investigation report.\\[4pt]
The Context is the factual layer of the investigation --- timeline, equipment, environmental conditions, operating procedures, anomalies, safeguards, and physical evidence. It must preserve the complexity of the case closely enough that a benchmark model can form its own investigative hypothesis from the material.\\[4pt]
Think like a benchmark constructor assembling the factual backbone of a case, not like a safety analyst writing conclusions. A strong Context feels like the factual record of an investigation where the meaning has not yet been assigned.\\[4pt]
\phead{Domain Policy}\\
- Use only the source investigation report. Do not use outside search.\\
- Respect the report's factual style and terminology.\\[4pt]
\phead{What to Include}\\
- Operating setting, facility/equipment descriptions, system configuration\\
- Full timeline with specific timestamps, readings, operator actions\\
- Environmental conditions and boundary parameters\\
- Operating procedures, maintenance history, inspection records\\
- Safeguards, alarms, interlocks, and their status\\
- Anomalies, warnings, prior incidents, or near-misses\\
- Physical evidence and post-incident observations\\
- Factual tensions: signals present but not acted upon, safeguards that did not prevent the outcome\\[4pt]
\phead{Leakage Policy} --- Must not contain: the agency's final probable cause; board conclusions or recommendations; post-hoc causal explanations; analysis-layer judgments about what ``caused'' or ``contributed to'' the incident.\\[4pt]
\phead{Self-Check}: (1) No causal language? (2) No probable cause/recommendations? (3) Detailed enough (timestamps, readings)? (4) Unresolved, not summarized? (5) Factual tensions preserved? (6) Evidentiary tone intact?\\[4pt]
\phead{Output}: \{``context'': ``<The benchmark context>''\}
\end{tcolorbox}

\subsubsection{Context Audit}

\begin{tcolorbox}[colback=gray!5, colframe=gray!40, coltitle=black, fonttitle=\small\bfseries, fontupper=\scriptsize\ttfamily, title={Context Audit --- Social Science}, breakable]
You are AuditAgent. You are reconstructing the pre-study reasoning behind a social science research paper.\\[4pt]
Read the source document and audit whether the draft Context is benchmark-ready as model-visible input. Do not rewrite the draft.\\[4pt]
\phead{What Benchmark-Ready Context Means}\\
- Long, formal, information-rich; carries enough background and unresolved tension to support a high-quality hypothesis.\\
- Must not leak answer-side contents: final conclusions, preferred method/mechanism, or post-hoc causal attribution.\\
- Time consistency: only information available before the hypothesis was formed.\\
- Should read like field or case background, not like paper narration or conclusion section.\\
- Should feel broader than a single paper's local framing.\\[4pt]
\phead{Hard Pass/Fail Gates}\\
1. Leaks answer-side content.\\
2. Does not go beyond the source paper's local framing.\\
3. Carries retrieval residue (citations, URLs, bibliography).\\
4. Too thin to support non-trivial hypothesis generation.\\[4pt]
\phead{Output}: \{``passed'': true\} or \{``passed'': false, ``summary'': ``...'', ``problems'': [...]\}
\end{tcolorbox}

\begin{tcolorbox}[colback=gray!5, colframe=gray!40, coltitle=black, fonttitle=\small\bfseries, fontupper=\scriptsize\ttfamily, title={Context Audit --- Safety Investigation}, breakable]
You are AuditAgent. Judge whether the draft Context is benchmark-ready model-visible input for a safety-domain hypothesis generation benchmark.\\[4pt]
\phead{What Benchmark-Ready Safety Context Means}\\
- Long, formal, information-rich; preserves the report's factual record and evidentiary tone.\\
- Contains timeline, equipment, environmental, operational, and anomaly details.\\
- Preserves factual tensions, competing signals, and unresolved questions.\\
- Does not leak the agency's probable cause, root-cause determination, or recommendations.\\[4pt]
\phead{Hard Pass/Fail Gates}\\
1. Leaks the agency's final probable cause or root-cause determination.\\
2. Contains analysis-layer causal language (``caused,'' ``contributed to'').\\
3. Reads like a summarized finding rather than a factual record.\\
4. Too thin to support non-trivial investigative hypothesis generation.\\[4pt]
\phead{Audit Philosophy}: Protect factual density, timeline fidelity, unresolved tension, and answer-side separation. Do not ask for better prose or methodology sections.\\[4pt]
\phead{Output}: \{``passed'': true\} or \{``passed'': false, ``summary'': ``...'', ``problems'': [...]\}
\end{tcolorbox}

\subsection{Stage 2: Hypothesis Construction}

\subsubsection{Hypothesis Forge}

\begin{tcolorbox}[colback=gray!5, colframe=gray!40, coltitle=black, fonttitle=\small\bfseries, fontupper=\scriptsize\ttfamily, title={Hypothesis Forge --- Social Science}, breakable]
You are ForgeAgent. You are reconstructing the pre-study reasoning behind a social science research paper.\\[4pt]
Your job is to reconstruct the single strongest benchmark-ready hypothesis that a careful researcher could state from the source document and the benchmark-visible Context, plus the evidence --- a research plan that explains how the hypothesis could be validated.\\[4pt]
\phead{Domain Guidance}\\
- Favor one behavioral, psychological, or organizational mechanism. Keep moderators, mediators, boundary conditions out unless one of them is the single central claim.\\
- The evidence plan should cover validation directions (replication, manipulation, mediation, moderation, boundary-condition, external-validity) in ex-ante language. Do not report statistics or study outcomes as already known.\\[4pt]
\phead{STEP 1 --- Formulate the Hypothesis}\\
Think like a benchmark constructor recovering the central bet worth stating before results were known.\\
- Distill into a single falsifiable claim in a short paragraph, using forward-looking language.\\
- Must be genuinely supportable from the Context alone.\\
- If the source supports several outcomes, compress into one higher-level central bet.\\[4pt]
\phead{STEP 2 --- Draft the Evidence}\\
A concise research plan as numbered items (one sentence each) describing validation directions --- comparisons, perturbations, mechanism probes, or boundary-condition tests. Should reflect the paper's actual evidence chain, not invent an ideal experiment from scratch.\\[4pt]
\phead{Self-Check}: (1) One claim or several bundled? (2) Context-supportable? (3) Forward-looking or summary? (4) Validation directions, not result recap?\\[4pt]
\phead{Output}: \{``hypothesis'': ``<...>'', ``evidence'': ``<...>''\}
\end{tcolorbox}

\begin{tcolorbox}[colback=gray!5, colframe=gray!40, coltitle=black, fonttitle=\small\bfseries, fontupper=\scriptsize\ttfamily, title={Hypothesis Forge --- Safety Investigation}, breakable]
You are ForgeAgent. Construct benchmark-ready hypothesis-evidence pairs from a safety investigation's model-visible Context.\\[4pt]
The source report contains both factual material and the agency's analysis/conclusions. You must read both, but hypotheses must be supportable from the Context alone.\\[4pt]
\phead{Core Rule: Same-Case Layering}\\
The source report contains a factual layer (what the Context preserves) and an analysis layer (causal explanations). Your hypotheses should be the ``explanatory middle ground'' --- judgments a careful investigator could form from the factual record, inspired by but not copied from the agency's analysis.\\[4pt]
\phead{Before You Begin}: Identify the agency's stated probable cause. Then deliberately construct hypotheses at a different level of specificity or from a different analytical angle.\\[4pt]
\phead{Hypothesis Style} --- Multiple pairs, each a different angle:\\
- accident mechanism or failure chain\\
- barrier failure or safeguard inadequacy\\
- human-system interaction or procedural gap\\
- maintenance, inspection, or design weakness\\
- organizational or management-system vulnerability\\[4pt]
\phead{Evidence Style} --- Each item must: point to a concrete context-visible fact; be one sentence; be factual rather than conclusory.\\[4pt]
\phead{Self-Check}: (1) Different analytical level from agency findings? (2) Distinct angles? (3) Context-supportable? (4) Investigative judgment, not restated finding? (5) Evidence points to concrete facts?\\[4pt]
\phead{Output}: \{``hypotheses'': [\{``hypothesis'': ``...'', ``evidence'': ``...''\}, ...]\}
\end{tcolorbox}

\subsection{Hypothesis Generation (Evaluation-Time Inference)}

\begin{tcolorbox}[colback=gray!5, colframe=gray!40, coltitle=black, fonttitle=\small\bfseries, fontupper=\scriptsize\ttfamily, title={Generation --- Social Science (Single-Hypothesis)}, breakable]
You are a social scientist examining existing theory and empirical findings.\\[4pt]
You are given a detailed background context from social science research. Your task is to formulate hypotheses.\\[4pt]
A strong hypothesis is:\\
- Evidence-bounded: grounded in the provided context, not speculative beyond the material\\
- Non-trivial: not an obvious restatement or surface-level observation\\
- Valuable: identifies a mechanism, process, or structural factor worth investigating\\
- Calibrated: claim strength proportionate to evidence support\\
- Specific: names concrete entities, processes, or relationships\\[4pt]
The hypothesis should be a single falsifiable or assessable claim as a short paragraph of plain prose. It should feel like a forward-looking judgment, not a summary.\\[4pt]
The evidence should be a research plan: numbered items describing how the hypothesis could be validated.\\[4pt]
Favor one claim about a behavioral mechanism, psychological process, or social phenomenon. The evidence plan should describe studies, comparisons, or measurement approaches.\\[4pt]
\phead{Output}: \{``hypotheses'': [\{``hypothesis'': ``<...>'', ``evidence'': ``<...>''\}]\}
\end{tcolorbox}

\begin{tcolorbox}[colback=gray!5, colframe=gray!40, coltitle=black, fonttitle=\small\bfseries, fontupper=\scriptsize\ttfamily, title={Generation --- Safety Investigation (Multi-Hypothesis)}, breakable]
You are a safety investigator examining the factual record of an incident.\\[4pt]
You are given a detailed context from safety investigation analysis. Your task is to produce multiple hypothesis-evidence pairs.\\[4pt]
A strong hypothesis is:\\
- Evidence-bounded: grounded in facts visible in the context, not speculative\\
- Non-trivial: goes beyond surface-level observations or obvious takeaways\\
- Valuable: identifies a mechanism, vulnerability, or intervention point\\
- Calibrated: claim strength matches evidence strength\\
- Specific: names concrete entities, processes, or relationships\\[4pt]
Each hypothesis should be a short paragraph identifying a distinct analytical angle. Each evidence section should contain numbered items --- concrete facts, timeline elements, conditions, or signals from the context.\\[4pt]
Focus on accident mechanisms, barrier failures, human-system interaction gaps, and organizational vulnerabilities.\\[4pt]
Each hypothesis must include a ``category'' field. Categories: ``causal'', ``latent'', ``intervention''.\\[4pt]
\phead{Output}: \{``hypotheses'': [\{``hypothesis'': ``...'', ``evidence'': ``...'', ``category'': ``causal|latent|intervention''\}, ...]\}
\end{tcolorbox}

\subsection{Arena Evaluation}

\begin{tcolorbox}[colback=gray!5, colframe=gray!40, coltitle=black, fonttitle=\small\bfseries, fontupper=\scriptsize\ttfamily, title={Arena Judge}, breakable]
You are an expert evaluator comparing two hypothesis outputs generated from the same context. Your goal is to identify which output demonstrates stronger evidence-grounded analytical reasoning.\\[4pt]
\phead{Judging Principles}\\
1. \phead{Length is not quality.} Evaluate density of well-grounded reasoning per claim. Extra length from speculative elaboration or hedging is a weakness.\\
2. \phead{Claims must be bounded by context evidence.} Naming entities from the context is restatement, not grounding. Naming entities NOT in the context is speculation.\\
3. \phead{Genuine synthesis vs.\ speculative elaboration.} Insight means integrating dispersed observations into a non-obvious explanatory structure the context supports but does not state.\\
4. \phead{Parsimony over proliferation.} Fewer well-grounded hypotheses beat many loosely grounded ones.\\
5. \phead{Calibration.} A strong claim with weak evidence is worse than a moderate claim with strong evidence.\\
6. \phead{Selectivity is expertise.} Developing the most important hypotheses demonstrates stronger judgment.\\[4pt]
\phead{Evaluation Process}\\
Important: Output A and Output B are in arbitrary order. Do not favor either based on position.\\[2pt]
Step 1 --- Read ALL hypotheses in BOTH outputs fully. Identify each output's single weakest hypothesis.\\
Step 2 --- Compare per quality dimension in 1--2 sentences.\\
Step 3 --- Determine verdict. The quality floor matters as much as the ceiling.\\[4pt]
\phead{Output}: \{``verdict'': ``A>>B'' | ``A>B'' | ``A=B'' | ``B>A'' | ``B>>A'', ``reason'': ``<one-sentence summary>''\}\\[2pt]
Use A=B only when genuinely comparable across all dimensions.
\end{tcolorbox}

\section{Skills}
\label{app:skills}

\small
\begin{longtable}{@{}>{\raggedright\arraybackslash}p{3cm} 
                   >{\raggedright\arraybackslash}p{3cm} 
                   >{\raggedright\arraybackslash}p{6.5cm}@{}}
\caption{Structured Analytical Skills for Hypothesis Generation} \label{tab:analytical_skills} \\
\toprule
\textbf{Skill Name} & \textbf{Category} & \textbf{Description} \\
\midrule
\endfirsthead

\multicolumn{3}{c}{{\bfseries \tablename\ \thetable{} -- Continued from previous page}} \\
\toprule
\textbf{Skill Name} & \textbf{Category} & \textbf{Description} \\
\midrule
\endhead

\midrule
\multicolumn{3}{r}{\textit{Continued on next page...}} \\
\endfoot

\bottomrule
\endlastfoot

\textbf{Baseline} & Single-pass Generation & 
Reads context and produces structured hypothesis sets (e.g., causal, latent, or analytical intervention-based for investigative domains; scientific research-oriented for academic domains, etc.) through evidence-first abductive reasoning. \\ \midrule

\textbf{ACH} (Analysis of Competing Hypotheses) & Evidence Matrix & 
Systematically evaluates multiple hypotheses using evidence-diagnosticity matrix with C/I/NA scoring. Ranks by inconsistency score (least disconfirming evidence wins). Guards against confirmation bias. \\ \midrule

\textbf{Chronology} & Temporal Analysis & 
Decomposes evidence into strict chronological sequence. Analyzes intervals, gaps, anomalous speeds, and temporal patterns to reveal hidden causal connections. Identifies critical transition points. \\ \midrule

\textbf{Assumptions Check} & Assumption Testing & 
Systematically identifies and stress-tests implicit assumptions underlying analysis. Categorizes by vulnerability (well-supported/reasonable/fragile). Generates alternatives from fragile assumptions. \\ \midrule

\textbf{Brainstorming} & Divergent-Convergent & 
Structured divergent-then-convergent process. Simulates multiple analytical perspectives (technical, human factors, organizational, contrarian). Generates broad candidate set before narrowing. \\ \midrule

\textbf{Cross-Impact Matrix} & Interaction Analysis & 
Examines how each factor interacts with every other factor using ++/+/0/--/--- notation. Reveals reinforcing loops, failed inhibitors, and overlooked connections. Identifies leverage points. \\ \midrule

\textbf{Devil's Advocacy} & Adversarial Challenge & 
Constructs strongest possible opposing case against leading hypothesis through evidence reliability, completeness, and reasoning attacks. Requires genuine commitment to opposing position. \\ \midrule

\textbf{Diagnostic Reasoning} & Evidence Quality & 
Identifies highest diagnostic value evidence---facts that most effectively discriminate between competing explanations. Prioritizes evidence quality over quantity. Builds hypotheses from diagnostic evidence outward. \\ \midrule

\textbf{Morphological Analysis} & Combinatorial & 
Decomposes problem into 3--5 key dimensions, enumerates possible values, systematically examines all combinations. Eliminates impossible, identifies high-plausibility and surprising combinations. \\ \midrule

\textbf{Premortem} & Backward Reasoning & 
Assumes leading hypothesis is WRONG, works backward to identify reasoning failures (misread evidence, missing evidence, alternative paths, overlooked factors). Analytical red-teaming. \\ \midrule

\textbf{Red Hat Analysis} & Actor Perspective & 
Adopts perspective of key actors---reconstructs their information state, pressures, training, incentives. Combats mirror imaging. Reasons how situation looks from their perspective, not analyst's. \\ \midrule

\textbf{Starbursting} & Dimensional (5W1H) & 
Generates hypotheses from six perspectives: What/When/Where/Who/How/Why. Performs cross-dimensional analysis to reveal interactions. Ensures comprehensive coverage without blind spots. \\ \midrule

\textbf{What-If Analysis} & Counterfactual & 
Assumes alternative outcome is proven true, works backward to construct causal path. Reframes from ``could this happen?'' to ``given it did happen, how?''. Breaks analytical anchoring. \\

\end{longtable}

\end{document}